\pdfoutput=1

\documentclass[11pt]{article}

\usepackage[final]{acl}

\usepackage{times}
\usepackage{latexsym}

\usepackage[T1]{fontenc}

\usepackage[utf8]{inputenc}

\usepackage{microtype}

\usepackage{inconsolata}

\usepackage{graphicx}
\usepackage{booktabs}
\usepackage{multirow}
\usepackage{fontawesome}
\usepackage{subcaption}
\usepackage{caption}
\usepackage{amsmath}
\usepackage{enumitem}
\usepackage{tablefootnote}
\usepackage[inkscapelatex=false]{svg}
\usepackage[textsize=scriptsize]{todonotes}

\newcommand{\ie}{\textit{i}.\textit{e}.}
\newcommand{\eg}{\textit{e}.\textit{g}.}

\newcommand{\Eg}{\textit{E}.\textit{g}.}
\newcommand{\etc}{\textit{etc}}

\usepackage{array}
\title{ImageInWords: Unlocking Hyper-Detailed Image Descriptions}

\author{Roopal Garg$^{1}$ \quad Andrea Burns$^{1}$ \quad Burcu Karagol Ayan$^{1}$ \\ {\bf Yonatan Bitton}$^{2}$ 
\quad {\bf Ceslee Montgomery}$^{1}$ \quad {\bf Yasumasa Onoe}$^{1}$ \quad {\bf Andrew Bunner}$^{1}$ \\
{\bf Ranjay Krishna}$^{3}$ \quad {\bf Jason Baldridge}$^{1}$ \quad {\bf Radu Soricut}$^{1}$
\\\\ 
\textsuperscript{1}Google DeepMind, \textsuperscript{2}Google Research, \textsuperscript{3}University of Washington\\ {\bf Data}: \url{https://github.com/google/imageinwords}\\ {\bf Correspondence}: \href{mailto:iiw-dataset@google.com}{iiw-dataset@google.com}
}
\begin{document}
\maketitle

\begin{abstract}
Despite the longstanding adage \textit{``an image is worth a thousand words,''} generating accurate hyper-detailed image descriptions remains unsolved. Trained on short web-scraped image-text, vision-language models often generate incomplete descriptions with visual inconsistencies. We address this via a novel data-centric approach with \textit{ImageInWords} (IIW), a carefully designed human-in-the-loop framework for curating hyper-detailed image descriptions. Human evaluations on IIW data show major gains compared to recent datasets (+66\%) and GPT-4V (+48\%) across comprehensiveness, specificity, hallucinations, and more. We also show that fine-tuning with IIW data improves these metrics by +31\% against models trained with prior work, even with only 9k samples. Lastly, we evaluate IIW models with text-to-image generation and vision-language reasoning tasks. Our generated descriptions result in the highest fidelity images, and boost compositional reasoning by up to $6\%$ on ARO, SVO-Probes, and Winoground datasets. We release the IIW-Eval benchmark with human judgement labels, object and image-level annotations from our framework, and existing image caption datasets enriched via IIW-model.
\end{abstract}

\section{Introduction}
\label{sec:intro}

Today's state-of-the-art Vision-Language Models (VLMs) are trained using large, noisy web datasets. WebImageText~\cite{radford2021learning}, ALIGN~\cite{jia2021scaling}, Conceptual Captions~\cite{sharma-etal-2018-conceptual} and LAION~\cite{schuhmann2022laion5b} rely on alt-text scraped from the internet as an imperfect image caption. Yet alt-text may only mention the photo location (\eg~``Europe''), the camera model used (\eg~``Canon EOS R6 Mark II''), or is SEO-specific (\eg,~``keep calm and carry on''). While data filtering and post-processing can remove noisy text, alt-text ambiguously captures image \textit{content} or \textit{intent}~\cite{enwiki:1189330128}. Therefore, only using image descriptions from the web is fundamentally flawed and limits model capabilities~\cite{thrush2022winoground,Shekhar_2017,Ma_2023_CVPR,ray2023cola,hsieh2024sugarcrepe}.

To curate better image-text data, recent work has released 
dense human written (DOCCI~\cite{docci}, DCI~\cite{urbanek2023picture}) or model generated caption datasets (PixLore~\cite{bonilla2023pixlore}, DAC~\cite{doveh2023dense}). Both have limitations, as using annotators without comprehensive guidelines results in outputs that vary by human attention, bias, and effort~\cite{burghardt2019quantifying,annburnout,PANDEY2022102772,ye2023cultural}. In contrast, model-generated captions are cheaper but incomplete and rife with hallucinations~\cite{rohrbach2019object,dai2023plausible}.

\begin{figure*}[tb]
\centering
\begin{subfigure}{\linewidth}
\centering
\includegraphics[width=\linewidth]{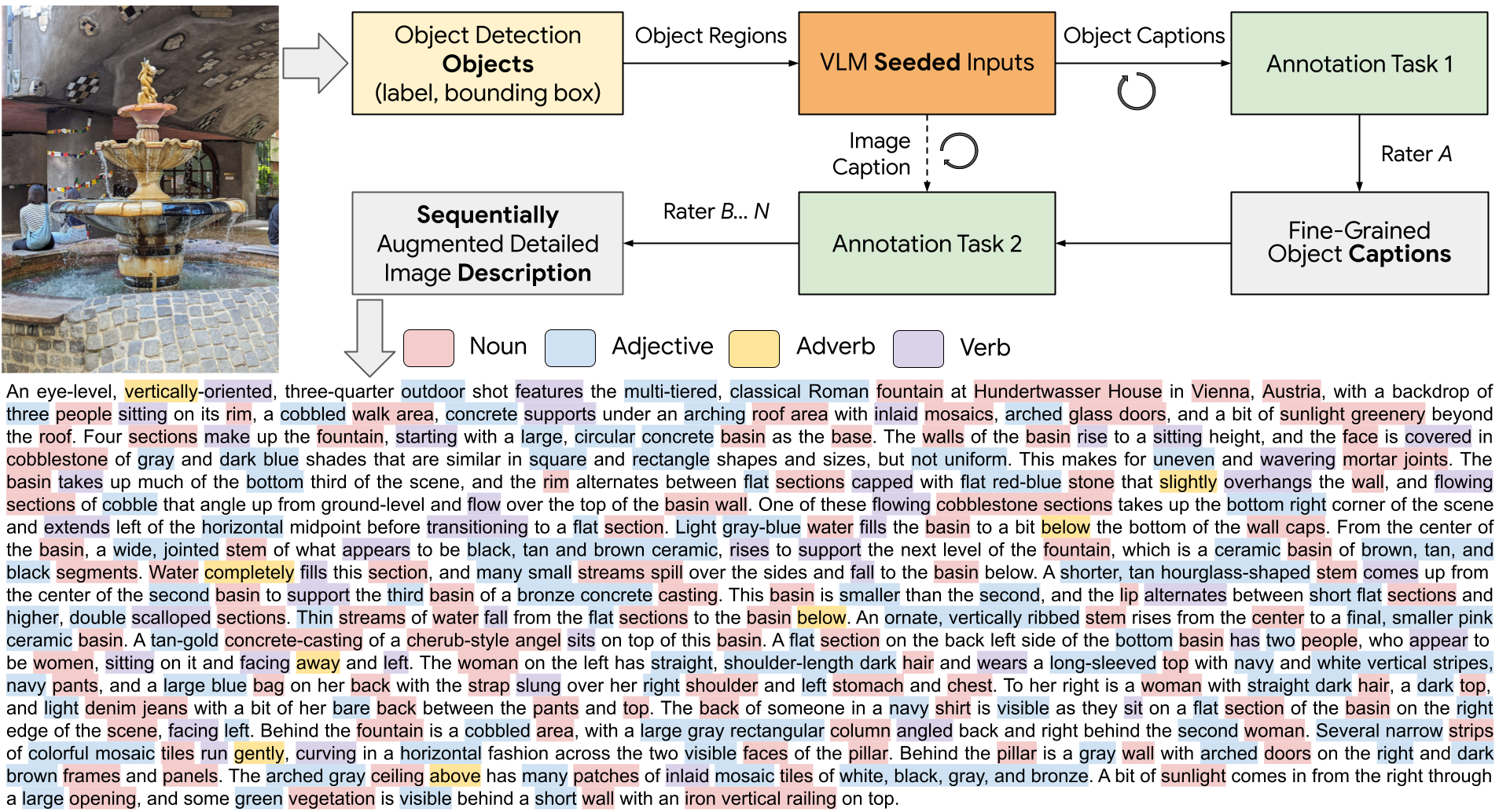}
\end{subfigure}
\caption{ImageInWords \textit{Seeded} Annotation Framework. Humans enrich and refine outputs \textit{sequentially}, building on prior human or machine inputs. Human annotation starts with fine-grained object captions in Task 1, which are used to compose image-level descriptions in Task 2. VLMs are updated in an active learning loop to produce better object and image-level seeds as annotated data becomes available. UI screenshots are in Appendix~\ref{subsec:uiexamples}.}
\label{fig:data-augmentation}
\vspace{-4mm}
\end{figure*}

In this work, we describe \textit{ImageInWords} (IIW), a human-in-the-loop framework for curating hyper-detailed image descriptions, and its resulting annotations. IIW combines the irreplaceable quality of human annotators with seeded metadata from machine generations. First, a VLM generates granular captions for each object in the image to seed our human annotation process, where crowd workers augment and fix the object-level captions to make them richer and hallucination free.

Next, at the image-level, a VLM generates a global caption to seed the final image \textit{description}. Crowd workers consume the image-level seed caption \textit{and} object-level human annotations to fill in contextual gaps. We design guidelines to attend to concepts beyond objects, such as visual perspective, spatial arrangement, and human object interactions. To ensure quality, multiple annotators iterate on a sample sequentially and we also incorporate active learning to produce better VLM seeds (Fig.~\ref{fig:data-augmentation}).

With this process, 
we construct the IIW dataset of 9018 hyper-detailed image descriptions. We find IIW has richer statistics than prior dense description datasets, with an average of 217.2 tokens, 52.5 nouns, 28 adjectives, 5 adverbs, and 19.1 verbs (Tab.~\ref{table:langcompare}). We assess quality with human side-by-side (SxS) comparisons to human-written datasets (DCI, DOCCI) and GPT-4V. Our descriptions are rated as more comprehensive, specific, human-like, with fewer hallucinations and better leading sentences at an average of +66\% (DCI, DOCCI) and +48\% (GPT-4V). 
We then fine-tune with IIW data and evaluate generated descriptions with the same SxS rubric: IIW model outputs are better by +31\% compared to models fine-tuned on prior work.

To better understand IIW models, we also perform text-to-image generation and vision-language reasoning experiments. Images generated with our model's descriptions are considered a closer reconstruction to the original image than when using other models. For vision-language compositionality, we replace images from ARO~\cite{yuksekgonul2023visionlanguage}, SVO-Probes~\cite{hendricks2021probing} and Winoground~\cite{thrush2022winoground} datasets with generated descriptions. IIW model descriptions help to better reason over attributes, relations, and word order compared to LLaVA-v1.5 and InstructBLIP descriptions.
\renewcommand{\arraystretch}{0.95}
\setlength{\tabcolsep}{4pt}
\begin{table*}[t!]
\centering
\begin{tabular}{l|c|c|c|c|c|c|c|c}
\hline
\multirow{2}{*}{Dataset}& Sample & Tokens & Tokens & Sentences & NN & ADJ & ADV & VB \\ 
\cline{3-9}
& Count & / Sentence & \multicolumn{6}{c}{/ Description}  \\
\hline 
SVP~\cite{krause2017hierarchical} & 19,561 & 11.9 & 68.5 & 5.7 & 17.1 & 6.7 & 1.1 & 5.0 \\
LocNar~\cite{PontTuset_eccv2020} & 873,107 & 15.7 & 41.0 & 2.6 & 10.7 & 1.6 & 0.4 & 3.5 \\
DCI$_{\text{extra}}$\tablefootnote{We use the extra\_caption field of DCI annotations and discuss this in choice in Section \ref{sec:background}. All following DCI references refer to the extra\_caption description.}~\cite{urbanek2023picture} & 7,805 & 15.8 & 148.0 & 9.3 & 35.3 & 16.3 & 3.6 & 10.5 \\
DOCCI~\cite{docci} & 14,647 & 19.2 & 135.7 & 7.1 & 34.0 & 16.6 & 2.7 & 9.6 \\
IIW (ours) & 9,018 & \textbf{22.1} & \textbf{217.2} & \textbf{9.8} & \textbf{52.5} & \textbf{28.0} & \textbf{5.0} & \textbf{19.1} \\
\hline 
\end{tabular}
\caption{Dataset Statistics Comparing ImageInWords (IIW) to Prior Work. We include the number of descriptions and the average number of tokens, sentences, nouns (NN), adjectives (ADJ), adverbs (ADV), and verbs (VB).}
\label{table:langcompare}
\vspace{-3mm}
\end{table*}

In summary, our contributions include:
\begin{itemize}[leftmargin=*]
\itemsep0em 
\item A human-in-the-loop annotation framework with extensive guidelines, iterative refinement, and VLM active learning that results in state-of-the-art hyper-detailed image descriptions.
\item Human SxS on comprehensiveness, specificity, hallucinations, human-likeness, and tldr-quality. Across these metrics, IIW data is better than recent DCI and DOCCI datasets by +66\% and +48\% better than GPT-4v, and +31\% better when used for fine-tuning than DCI and DOCCI.
\item IIW model evaluations with text-to-image generation and vision-language compositional reasoning tasks to complement human SxS. IIW model descriptions generate images most similar to the original image (ranked 1st) and improve distinguishing true image-text pairs given attribute, relation, or word order differences by up to 6\%.
\item An open source IIW-Eval benchmark of human and model annotations over 2.6k images and their image descriptions, and 1.9k object descriptions. We also release human SxS labels between IIW, DCI, and DOCCI for comparison in future work.
\end{itemize}
\section{Related Work}
\label{sec:background}

Image \textit{captioning} has been studied for years, starting with CNN and LSTM encoder-decoder frameworks for generic captions~\cite{vinyals2015tell,anderson2018bottomup}, to the more recent Transformer-based VLMs for more difficult captions~\cite{chen2023pali,li2023blip2} (\eg, VizWiz~\cite{gurari2020captioning}, NoCaps~\cite{agrawal2019nocaps}, TextCaps~\cite{sidorov2020textcaps}). These datasets and many others contain captions with 15 words or fewer~\cite{desai2021redcaps,flickr30k,lin2015microsoft,mao2016generation,plummer2015flickr30k,kazemzadeh2014referitgame,krishna2016visual,plummer2015flickr30k} and may differ by caption grounding level (\eg whole image or region-level captions) or image domain (\eg images taken by people who are blind or images capturing text).

However, few \textit{dense} image \textit{description} datasets exist. PixLore~\cite{bonilla2023pixlore} used multiple vision-language datasets to generate verbose captions with BLIP-2~\cite{li2023blip2}. DAC~\cite{doveh2023dense} uses a machine-generated approach: pretrained LLMs expand the original image caption and pretrained VLMs generate captions over smaller image regions. The resulting descriptions are used to fine-tune a VLM model for better compositional reasoning. While model-only approaches are cost effective and avoid the challenges of designing annotation instructions, they risk introducing hallucinations and systematic biases.

DOCCI~\cite{docci} collects image descriptions with only crowd workers, which we later show can be considerably improved. Closest to IIW is DCI~\cite{urbanek2023picture}, which uses human annotators to reach denser descriptions. DCI uses the SAM~\cite{kirillov2023segment} object detector to generate smaller regions to be described and then composes them into an overall description.

DCI's available annotations and metadata can be concatenated with additional text to reach 1k+ length. However, filler text and image labels are used to reach this length, and repeated or highly overlapping sentences are often present. As a result, we use their ``extra\_caption'' field for fair comparison as it is the only coherent description available. In contrast to DCI, we also allow crowd workers to update or correct every component of the seeded information. IIW output is then sequentially refined over multiple annotation rounds to produce a single coherent annotation. In comparison to DCI's ``extra\_caption'' annotation, we collect significantly better descriptions, as reflected in Tab.~\ref{table:langcompare} statistics.
\section{ImageInWords Dataset Collection}
\label{sec:dataset}
\begin{figure*}[t!]
\centering
\begin{subfigure}{0.245\linewidth}
\includegraphics[page=1, width=\linewidth]{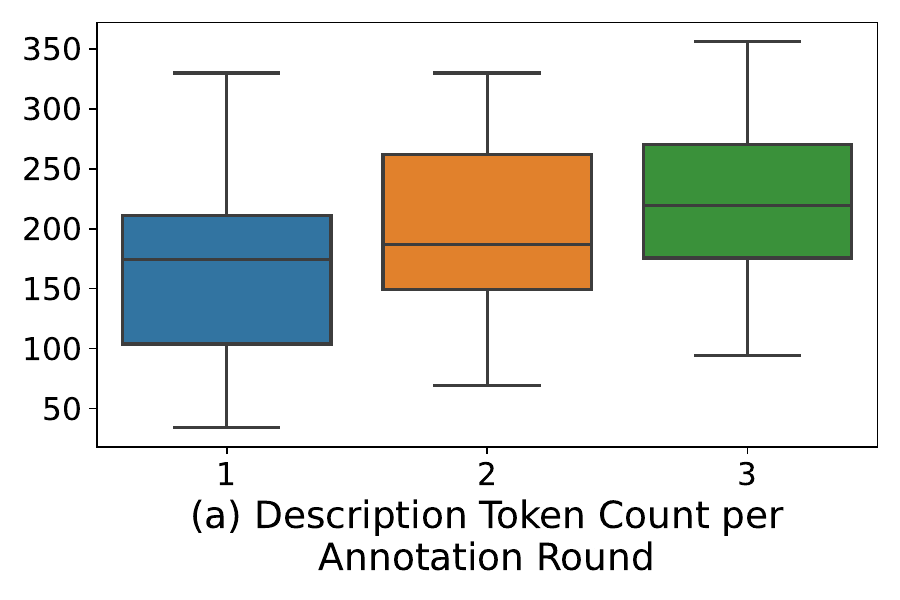}
\end{subfigure}
\begin{subfigure}{0.245\linewidth}
\includegraphics[page=3, width=\linewidth]{figures/sequential_annotation_initial_rounds.pdf}
\end{subfigure}
\begin{subfigure}{0.245\linewidth}
\includegraphics[page=2, width=\linewidth]{figures/sequential_annotation_initial_rounds.pdf}
\end{subfigure}
\begin{subfigure}{0.245\linewidth}
\includegraphics[width=\linewidth]{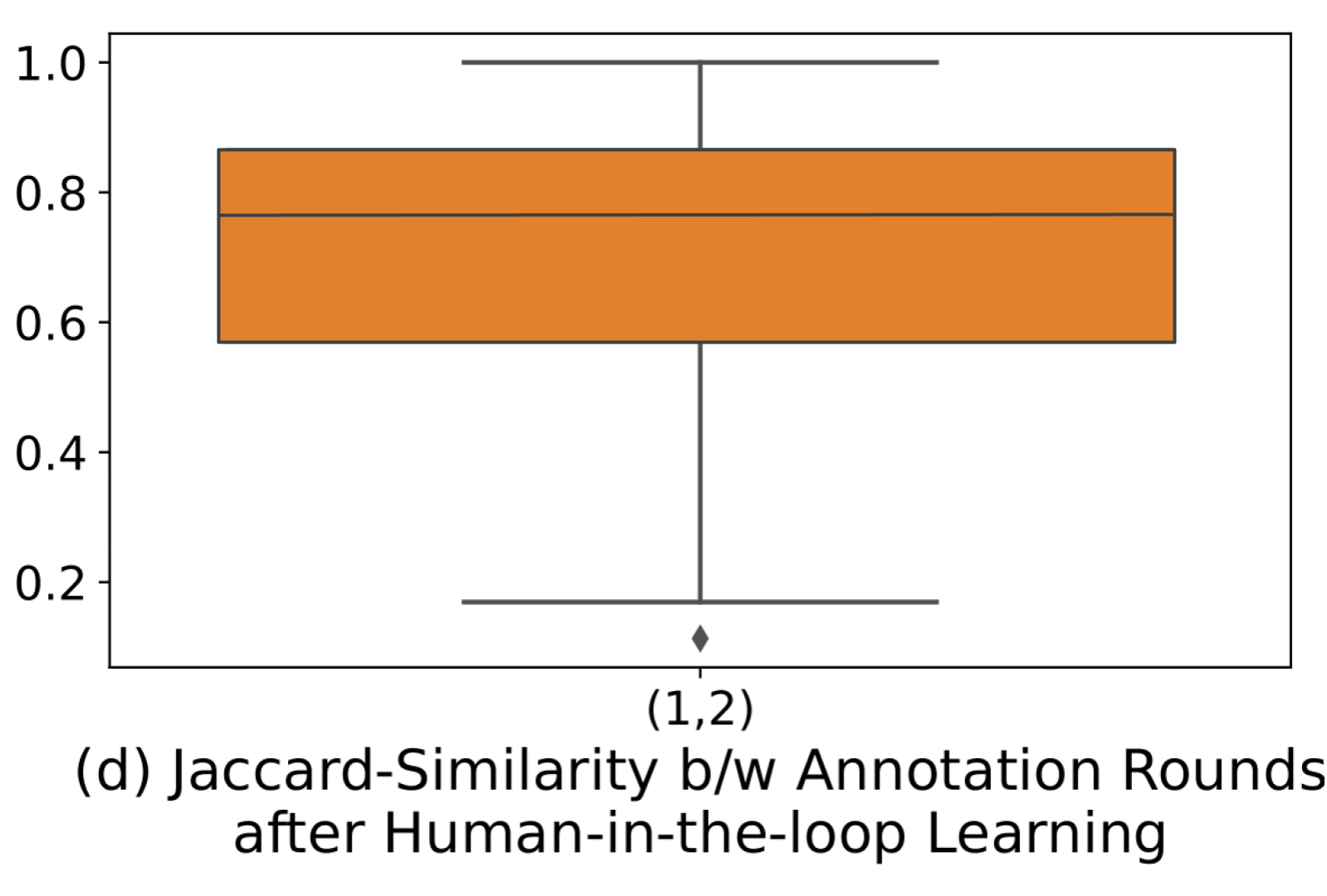}
\end{subfigure}
\caption{Effects of Sequential Annotation: Over annotation rounds, (a) token count goes up as (b) time spent goes down with (c) higher agreement, measured by Jaccard Similarity~\cite{enwiki:1196092673}. (d) Over time with a constant human annotator pool, each learns from the other via an \textit{implicit feedback loop} and a high agreement rate in round (1,2) can now be observed as was previously only seen in round (2,3) in (c).}
\label{fig:sequential_annotation_rounds}
\vspace{-3mm}
\end{figure*}

The IIW dataset is composed of 9018 (Train: 8573, Test: 445) images that are sampled from a WebLI~\cite{chen2023pali} like dataset and human annotated. Details on the human annotator pool are provided in Appendix~\ref{subsec:annpool}. In~\ref{sec:guidelines}, we briefly review our foundational guidelines for crowd workers. Annotation methodology and the types of image-text annotations we collect are described in~\ref{sec:methodology} and~\ref{sec:setup}.

\subsection{Annotation Guidelines}
\label{sec:guidelines}

We compile an extensive set of guidelines for human annotators and iterate over them with multiple pilot rounds. Appendix~\ref{sec:supplemental-guidelines} contains the complete set of guidelines due to space. Annotators are asked to only include details that can be deduced from visual cues, erring on the side of higher precision. To compose coherent descriptions, unnecessary fragmentation of sentences and the use of filler phrases like \textit{``in this image,}'' ``\textit{we can see},'' and \textit{``there is a,''} should be avoided since they add no visual detail.

While describing the overall image, we instruct annotators to start with a newspaper style TLDR (Too Long Didn't Read; meant to serve as a succinct summary). Objects should be described in the order of their saliency, noting objects and relationships in a well organized manner. Descriptions should include the overall setting, background, and style, considering the camera angle, overall composition, and rendered text. We also ask to pay special attention to people, apparel, art pieces, locale-specific, and unique attributes with the following as example features: function, shape, size, color, design, pattern, texture, material, condition, opacity, orientation, location, relationship to other components/objects, and text written on objects.

\subsection{Annotation Methodology}
\label{sec:methodology}
This section describes the \textit{seeded}, \textit{sequential} process employed in annotating the IIW dataset. We highlight that IIW data is meant for supervised fine-tuning rather than pretraining. As a result, our goal was to annotate a small-scale, high quality dataset.
Still, we designed the human-in-the-loop process to be as efficient and flexible as possible. The number of sequential annotators and the presence of Task 1 can be adjusted as time and budget permit.

\noindent\textbf{Seeded Annotation}
\label{sec:seeded}
Describing images in detail is highly subjective and complicated. To expedite human annotation, we use PaLI-3 5B outputs to seed the annotation process instead of crowd workers starting from scratch. While VLMs have improved in their ability to capture image details, attempts to generate a consistent rich output still fall prey to hallucinations and recall issues. Our human annotation pipeline ensures that VLM hallucinations can be corrected and missing details filled in.

An initial machine generated caption and high precision, domain specific metadata (\eg, art style or title of a painting) provide a minimal quality and coverage guarantee. As data is collected, the VLMs used for seeding are updated to produce better quality descriptions in an active learning loop
(reflected with loops in Figure~\ref{fig:data-augmentation}). After batches of 1k samples are annotated, we retrain (i.e., re-fine-tune) the PaLI-3 5B models with all available annotations (for both Task 1 and Task 2). 

We find that these updates significantly improve the baseline model, with early batches shifting PaLI captions from an average of 15 to 150+ words with as few as 3k samples. We do not yet perform specialized sampling for active learning due to the large performance gap between the ImageInWords human annotations and ImageInWords model (as later shown in Tab.~\ref{tab:human-vs-model-iiw}). However, this could be incorporated in the future if performance saturates.

\noindent\textbf{Sequential Augmentation}
\label{sec:sequential}
We further improve framework efficiency with sequential description augmentations. Humans augment a previous crowd worker's and/or VLM's outputs instead of starting from scratch. After the first augmentation, both the machine-generated seed and prior human annotation are provided. The following annotators do not know which is model output versus human written, which can mitigate preference to model outputs.

During the annotation process, it is far more effective in \textit{time} and \textit{quality} to read and augment image descriptions: 
in Fig.~\ref{fig:sequential_annotation_rounds} we see that if annotations were done in parallel, we would have 3 competing outputs per image, each with their own style, perspective, and weaknesses, with each containing $\sim$170 words and taking $\sim$800 seconds. Whereas, in the sequential process, we get a single all-inclusive description that has been verified and augmented by three humans with +20\% token count in -30\% time. Higher Jaccard similarity over rounds suggests a higher inter-annotator agreement, which also serves as a proxy for quality.

Finally, our framework has an implicit human-to-human learning loop, as each human annotator has the opportunity to read and learn from other perspectives across the annotation rounds, leading to improved individual quality. 
This is seen in the $\sim$2x improved inter-annotator agreement between rounds (1, 2) when comparing (c) and (d) in Fig.~\ref{fig:sequential_annotation_rounds}.

\begin{figure*}[t!]
\centering
\begin{subfigure}{\linewidth}
\centering
\includegraphics[width=\linewidth]{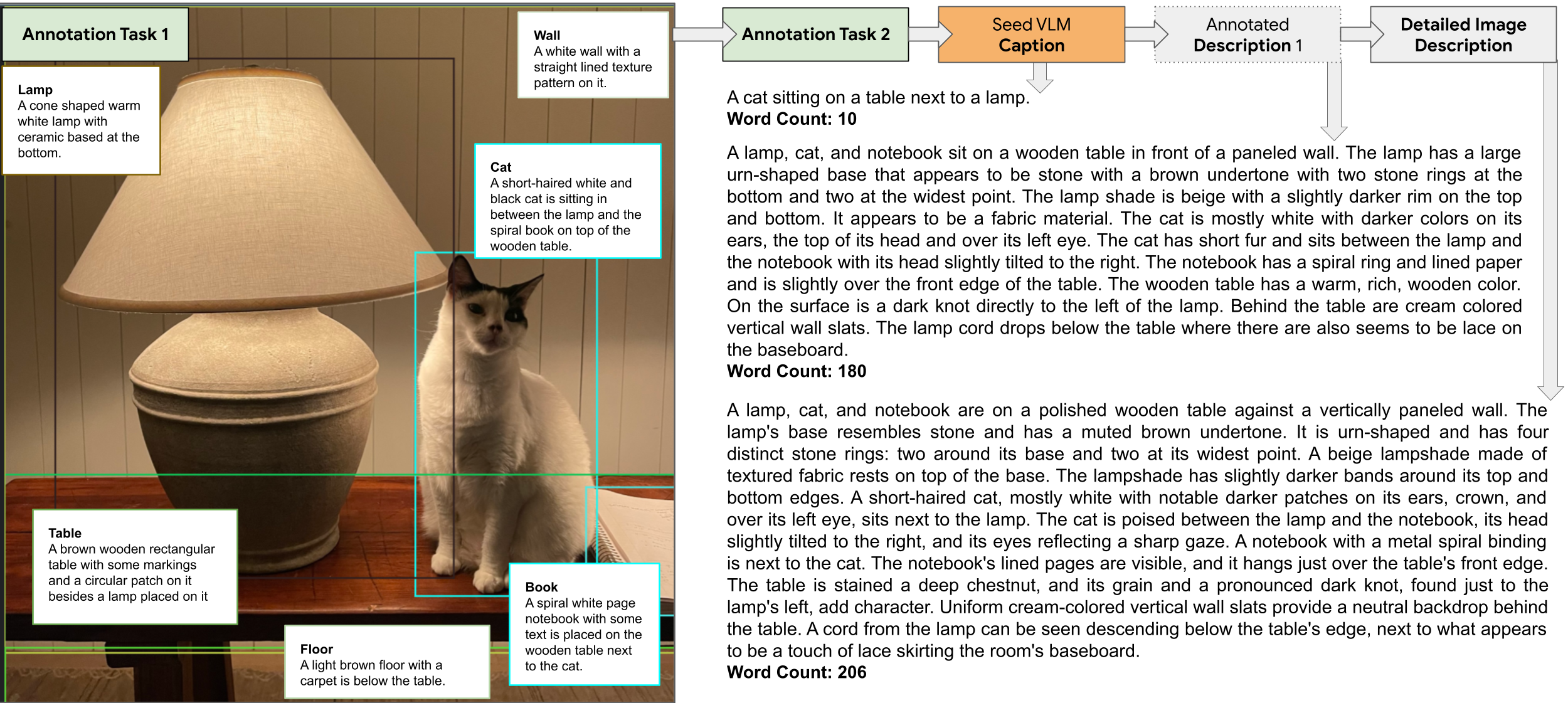}
\end{subfigure}
\caption{IIW Annotation Tasks. Objects and their attributes are first individually annotated to note the salient objects and focus on coverage of their attributes in Task 1. These outputs, along with a seed VLM caption, are passed to humans to build the initial image-level description. The initial caption is then human augmented and refined in \textit{N} sequential rounds to attain the final hyper-detailed description in Task 2.}
\label{fig:data-augmentation-example}
\vspace{-3mm}
\end{figure*}

\subsection{Annotation Framework}
\label{sec:setup}

Based on the above guidelines, we present the IIW framework for annotating images across two tasks. The tasks are seeded from VLMs or prior human annotations (Fig.~\ref{fig:data-augmentation-example}), where each can have multiple annotation rounds. Examples are in Appendix~\ref{subsec:uiexamples}.

\vspace{2mm}

\noindent\textbf{Task 1: Object-Level Descriptions}
Similar to Visual Genome~\cite{krishna2016visual}, we design this annotation task to capture a (label, bounding box, object description) triplet per salient image object. An object's label is open vocabulary with no verbosity restrictions, and its description is focused on the object but additionally takes the context of the image into account. The bounding box localizes where the object is in the image (Fig.~\ref{fig:data-augmentation-example} (left)). To seed the data, we first used an internal object detection (OD) model to obtain a list of (label, bounding box) pairs. Then, object captions are generated by cropping the image to the object bounding box and generating a caption via a periodically fine-tuned PaLI-3 5B. Our methodology is agnostic to which VLM, OD (or image-segmentation) model is used.

From the seed list of (label, bounding box, object caption), the annotators are first asked to determine the salient objects and fix the list of (label, bounding box) by editing, removing, adding or merging the object annotations based on their accuracy, importance, and role in the overall image. By limiting the scope to individual objects, annotators can better focus and capture details comprehensively. 

\vspace{2mm}

\noindent\textbf{Task 2: Image-Level Descriptions} 
Our second annotation task is to form the final hyper-detailed description. Task-1 outputs, optional domain specific metadata (\eg, art style of a painting), and a VLM seed caption are used to hint and help the annotators compose the overall image description.

The bulk of the annotation responsibility falls on the first annotator; note that crowd worker annotation order is randomly assigned per sample and the same annotator is not re-employed for the same sample. This output is then refined and augmented in sequential rounds to mitigate subjectivity and quality drops. Annotators are encouraged to focus on augmentation and only remove things if they are obvious errors, but are free to re-frame information to add new details. We started with 3 annotation rounds and monitored the n-gram Jaccard similarity between the outputs. Once a 0.8 round-over-round output similarity was achieved, we reduced the numbers of rounds. Optionally, early stopping support could be added to the annotation framework itself to make this instance specific. Over time, we found our similarity threshold can be met between the first two rounds, \ie, (1,2), (Fig.~\ref{fig:sequential_annotation_rounds}) suggesting improved and high individual-annotator quality.
\section{IIW Human-Authored Data Eval}
\label{sec:data_experiments}

To evaluate the IIW annotation framework and resulting human annotations, we  start with human SxS evaluations to compare our human annotations to prior work (\eg DCI, DOCCI, GPT-4V). To run a SxS experiment on human-authored description quality, we first need a common pool of human annotated images. For this, we additionally annotate the DCI test set (112) and a comparable number of samples (100) from the DOCCI test set with our IIW annotation framework. We thus have human-authored IIW annotations for direct comparison on images in the DCI and DOCCI datasets, which contribute to our open-source IIW-Eval benchmark.

Our human SxS framework evaluates 5 metrics: Comprehensiveness, Specificity, Hallucinations, quality of the first few line(s) as a TLDR (Too Long Didn't Read; meant to serve as a succinct summary), and Human-Likeness. Comprehensiveness concerns whether a description covers all key information and objects present in an image. Specificity is the degree of detail in which each of the key objects and details are described in.

We also include TLDR quality as one of our metrics as initial sentences set a precedence for what details to expect, both for the reader and models trained on this data. From a practical perspective, we would like hyper-detailed descriptions to still be useful in a setting that is constrained by input text length; \ie, if we truncate an image description, it should contain the most salient information for vision-language training.
While IIW guidelines instruct annotators to include a first sentence which provides an overall summary of the image content, prior work also designed their descriptions to start with either a short caption that summarizes the full image \cite{urbanek2023picture} or have important information covered in earlier sentences \cite{docci}. As a result, we believe the TLDR metric is reasonable and should be an established practice for hyper-detailed descriptions moving forward.

The evaluation is done on a 5 point scale defined using ``substantially better'' (\text{\tiny\faPlus\space\faPlus}) or ``marginally better'' (\text{\tiny\faPlus}) ratings on both sides of a ``neutral'' (-). Higher numbers indicate higher quality across each metric, and our tables report \textit{percentages} for ease of comparison. We emphasize that this is an \textit{extremely challenging} human annotation task, where per image, two text pieces of 100+ words need to be evaluated across 5 metrics in a SxS setting. On average, we observe each comparison takes 15-20 minutes.
Details on the annotation setup and UI are in Appendix~\ref{subsec:uiexamples}.
\renewcommand{\arraystretch}{0.92}
\begin{table}[t!]
\centering
\begin{tabular}{p{0.8cm}|ccccc|ccccc}
\hline
\multirow{3}{*}{\rotatebox[origin=c]{315}{Metric}} & \multicolumn{5}{c|}{DCI Test}&\multicolumn{5}{c}{DOCCI Test}\\
\cline{2-11}
& \multicolumn{2}{c|}{DCI}&&\multicolumn{2}{|c|}{IIW}& \multicolumn{2}{c|}{DOCCI}&& \multicolumn{2}{|c}{IIW}\\
\cline{2-11}
& \text{\tiny\faPlus\faPlus} & \text{\tiny\faPlus} & - & \text{\tiny\faPlus}&  \text{\tiny\faPlus\faPlus} &  \text{\tiny\faPlus\faPlus} & \text{\tiny\faPlus}& - & \text{\tiny\faPlus}&  \text{\tiny\faPlus\faPlus}\\
\hline
C & 3 & 7 & 19 & 30 & \textbf{41} & 4 & 6 & \textbf{38} & 33 & 19 \\
S & 5 & 3 & 4 & 20 & \textbf{68} & 3 & 2 & 8 & 22 & \textbf{65} \\
H & 2 & 3 & \textbf{48} & 32 & 15 & 0 & 12 & \textbf{41} & 34 & 13 \\
Tldr & 3 & 0 & 3 & 20 & \textbf{74} & 1 & 4 & 11 & 30 & \textbf{54} \\
HL & 1 & 1 & 14 & 25 & \textbf{59} & 1 & 0 & 30 & \textbf{46} & 23 \\
\hline
\end{tabular}
\caption{Human SxS to Evaluate IIW Human-Authored Data. We report percentages comparing data from prior work with data annotated by the IIW framework on Comprehensiveness (C), Specificity (S), Hallucinations (H), TLDR-quality, and Human-Likeness (HL).}
\label{table:sxs-human}
\vspace{-3mm}
\end{table}

\subsection{Human SxS Results}
\label{subsec:datahumaneval}
\renewcommand{\arraystretch}{0.92}
\setlength{\tabcolsep}{3.35pt}
\begin{table*}[t!]
\centering
\begin{tabular}{l|ccccc|ccccc|ccccc|ccccc}
\hline
\multirow{4}{*}{Metric} & \multicolumn{15}{c|}{\textcolor{blue}{Model} Generated} &  \multicolumn{5}{c}{\textcolor{blue}{Model}-\textcolor{red}{Human}}\\
\cline{2-21}
& \multicolumn{10}{c|}{LocNar Eval} & \multicolumn{5}{c|}{IIW-Eval}& \multicolumn{5}{c}{IIW-Eval} \\
\cline{2-21}
& \multicolumn{2}{c|}{\textcolor{blue}{DCI}} & & \multicolumn{2}{|c|}{\textcolor{blue}{IIW}} &  \multicolumn{2}{c|}{\textcolor{blue}{DOCCI}} & & \multicolumn{2}{|c|}{\textcolor{blue}{IIW}} &
\multicolumn{2}{c|}{\textcolor{blue}{GPT-4V}}& & \multicolumn{2}{|c|}{\textcolor{blue}{IIW}} &
\multicolumn{2}{c|}{\textcolor{blue}{GPT-4V}}& & \multicolumn{2}{|c}{\textcolor{red}{IIW}} \\
\cline{2-21}
& \text{\tiny\faPlus\faPlus} & \text{\tiny\faPlus} & - & \text{\tiny\faPlus}&  \text{\tiny\faPlus\faPlus}& \text{\tiny\faPlus\faPlus} & \text{\tiny\faPlus}  & - & \text{\tiny\faPlus}&  \text{\tiny\faPlus\faPlus}& \text{\tiny\faPlus\faPlus} & \text{\tiny\faPlus}  & - & \text{\tiny\faPlus}&  \text{\tiny\faPlus\faPlus} &\text{\tiny\faPlus\faPlus} & \text{\tiny\faPlus} & - & \text{\tiny\faPlus}&  \text{\tiny\faPlus\faPlus}\\
\hline
Comprehensive & 7 & 10 & 24 & \textbf{32} & 27 & 5 & 22 & \textbf{42} & 26 & 5 & 21 & 29 & \textbf{36} & 10 & 4 & 3 & 10 & \textbf{39} & 29 & 19\\
Specificity & 6 & 10 & 14 & 24 & \textbf{46} & 6 & 14 & 23 & \textbf{33} & 24 & \textbf{46} & 32 & 12 & 8 & 2 & 6 & 10 & 15 & \textbf{35} & 34 \\
Hallucinations & 12 & 21 & \textbf{43} & 11 & 13 & 9 & 25 & \textbf{39} & 21 & 6 & 22 & \textbf{29} & 23 & 20 & 6 & 0 & 6 & 29 & \textbf{34} & 31\\
TLDR & 9 & 11 & 9 & 30 & \textbf{41} & 6 & 7 & 17 & \textbf{42} & 28 & 7 & 15 & 27 & \textbf{31} & 20 & 5 & 6 & 8 & \textbf{47} & 34\\
Human-Like & 11 & 5 & 13 & 32 & \textbf{39} & 6 & 12 & \textbf{41} & 27 & 14 & 8 & 22 & \textbf{60} & 7 & 3 & 6 & 13 & \textbf{41} & 27 & 13\\
\hline
\end{tabular}
\caption{Human SxS on Model Predictions. Model Generated compares PaLI-5B fine-tuned with IIW versus prior work DCI and DOCCI and GPT-4V outputs. Model-Human compares GPT-4V model to IIW human-annotations.}
\label{table:sxs-human-locnar}
\vspace{-2mm}
\end{table*}

Tab.~\ref{table:sxs-human} reports preference percentages for each human-authored test set on our five metrics. Comparing IIW to DCI and DOCCI, Comprehensiveness is higher by +61\% and +42\%, Specificity by +80\% and +82\%, Hallucinations are lower by 42\% and 35\%, TLDR quality is higher by +91\% and +79\%, and Human-Likeness improves by +82\% and +68\%, respectively. This indicates that the IIW human-authored image descriptions on images from DCI and DOCCI are considerably better than those originally published with prior work.

To further quantify the quality of IIW human annotations, we compare with GPT-4V outputs~\cite{openAI2023gpt4v} in Tab.~\ref{table:sxs-human-locnar} (right). We use GPT-4V to generate image descriptions on 100 IIW-Eval images. The descriptions are generated with the prompt ``Generate a detailed image description'' and no other specifications. The results from the Model-Human section of Tab.~\ref{table:sxs-human-locnar} show that we reach Comprehensiveness (+35\%), Specificity (+53\%), Hallucination (+59\%), TLDR (+70\%), and Human-Likeness (+21\%) improvements over GPT-4V outputs. Although GPT-4V performs relatively better than the human-authored DCI and DOCCI data when compared to IIW annotations, we assess that considerable future modeling efforts are needed for VLMs to reach IIW human-authored data quality.

\section{IIW Model Evaluation}
\label{sec:model_experiments}

After evaluating IIW human annotations, we turn to quantifying the impact of fine-tuning with IIW data versus fine-tuning with prior work. We fine-tune separate PaLI-3 5B models on DCI, DOCCI and IIW training splits, with their detailed human-authored text as target. Each model is trained with an identical setup ($\sim$40 epochs, learning rate 3e-4, batch size 32) and the generic input instruction: ``Generate a detailed image description.'' More fine-tuning details are provided in Appendix~\ref{sec:supplemental-tasks} and~\ref{sec:supplemental-experiments}.

As shown in prior work, existing text similarity metrics like BLEU~\cite{papineni-etal-2002-bleu} and ROUGE~\cite{lin-2004-rouge} have been shown to poorly correlate with human judgement as they are heavily dependent on n-gram overlaps, and thus ill-suited for long texts~\cite{kryscinski2019neural,caglayan-etal-2020-curious}. Prior works DAC, DCI, and DOCCI also are limited by existing image caption metrics, and use LLM summaries of their descriptions or human SxS for evaluation. We report BLEU, ROUGE, CIDEr, BERTScore~\cite{bertscore}, and BLEURT~\cite{pu2021learning} in Appendix~\ref{sec:textsimmetrics} but look to human SxS for more accurate judgements. 

We also quantify the richness of the IIW model outputs via two downstream evaluations which can help us to evaluate IIW model generated descriptions in the absence of better metrics. First, in~\ref{subsec:t2i}, we use generated descriptions from DCI, DOCCI, and IIW fine-tuned models to prompt a Text-to-Image (T2I) model for image reconstruction and evaluate which descriptions result in higher fidelity generated images. Then, in~\ref{subsec:llmvqa}, we quantitatively show how IIW models can generate descriptions to aid in vision-language reasoning. 

\subsection{Human SxS Results}
\label{subsec:modelhumaneval}

Our first evaluation uses the same human SxS setup as in Section~\ref{sec:data_experiments}.
We evaluate the IIW, DCI, and DOCCI fine-tuned models on a random sample of LocNar Eval images, which can serve as an unseen test set for each fine-tuning dataset. The results mirror Tab.~\ref{table:sxs-human}'s human-authored statistics: IIW has gains over (DCI, DOCCI) datasets on Comprehensiveness (+42, +4)\%, Specificity (+54, +37)\%, TLDR (+51, +57)\% and Human-Likeness (+55, +23)\% with a relatively small hallucination trade-off (-9, -7)\%, largely dominated by marginal rated losses. Overall, compared to DCI and DOCCI, IIW model-generated outputs show a higher average preference from human judgement by +31\%.

From Tab.~\ref{table:sxs-human-locnar} (middle), we see that the IIW PaLI-5B fine-tuned model has clear room for improvement compared to GPT-4V, as expected given its 5B size. It is worth noting that it competes well on the Human-Likeness writing-style metric, and actually excels at learning the TLDR concept, which we built as a distinct feature of our dataset.

\begin{figure*}[t!]
\centering
\begin{subfigure}{\linewidth}
\centering
\includegraphics[width=\linewidth]{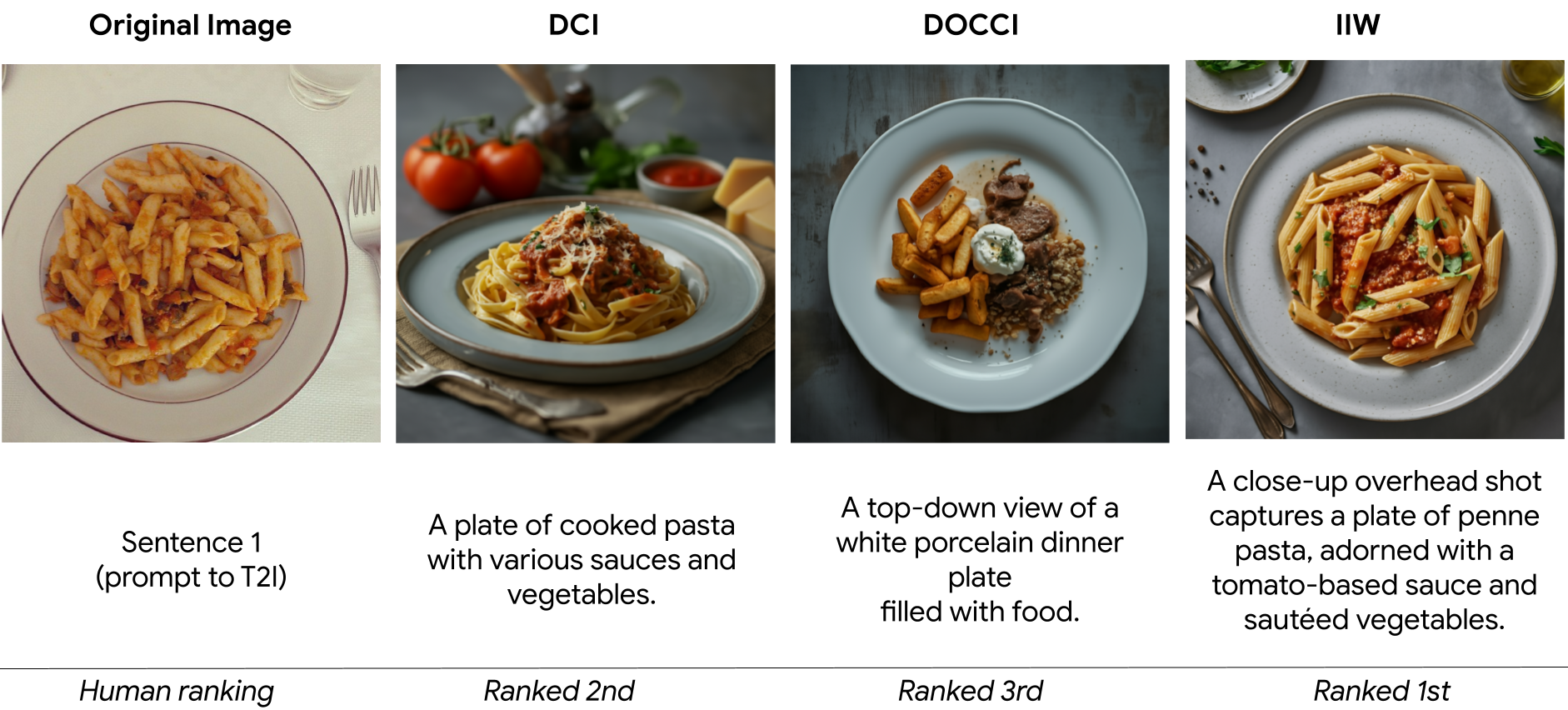}
\label{fig:ImageInWords_Figures-reconstruction}
\vspace{-3mm}
\end{subfigure}
\caption{Example T2I Outputs and Human Rankings. We show an example output when the first sentence of the image description from DCI, DOCCI and IIW PaLI-5B fine-tuned models are fed as input to the same T2I model.}
\label{fig:reconstruction}
\end{figure*}

\setlength{\tabcolsep}{7pt}
\begin{table}[t!]
\centering
\begin{tabular}{l|ccccc}
\hline 
\multirow{2}{*}{PaLI-ft}  & \multicolumn{5}{c}{Mean Rank $\downarrow$}\\
\cline{2-6}
& 1 & 1-2 & 1-3 & 1-4 & 1-5 \\
\hline
DCI & 2.05 & 2.06 & 1.95 & 2.00 & 1.88 \\
DOCCI & 1.74 & 1.79 & 1.83 & 1.84 & 1.86 \\
IIW & \textbf{1.63} & \textbf{1.69} & \textbf{1.62} & \textbf{1.66} & \textbf{1.66} \\
\hline
\end{tabular}
\newline
\vspace*{3mm}
\newline
\setlength{\tabcolsep}{8pt}
\begin{tabular}{l|cccc}
\hline 
\multirow{2}{*}{PaLI-ft} & \multicolumn{4}{c}{CLIP Image Similarity $\uparrow$}\\
\cline{2-5}
& 1 & 1-2 & 1-3 & 1-4 \\
\hline
DCI & 0.844 & 0.852 & 0.855 & 0.850 \\
DOCCI & 0.853 & 0.862 & 0.865 & 0.855 \\
IIW & \textbf{0.861} & \textbf{0.867} & \textbf{0.870} & \textbf{0.868} \\
\hline
\end{tabular}
\caption{T2I Reconstruction from Image Descriptions. The original image is compared to images generated from cumulative sentence inputs on relative (Mean Rank) and absolute (CLIP image similarity) metrics.}
\label{table:reconstruction-results}
\vspace{-4mm}
\end{table} 
\subsection{Reconstructing Images with IIW}
\label{subsec:t2i}

To complement our SxS analysis, we consider how IIW generated descriptions can empower T2I models to produce more controlled and specific image reconstructions. For this study, we use the PaLI-5B (DCI, DOCCI and IIW) fine-tuned VLMs to generate descriptions on 240 images from the LocNar eval set. We then split each image description into sentences as units which are fed as cumulative inputs (\ie, sentence 1, sentence 1-2, sentence 1-3...) to an Imagen model variant~\cite{saharia2022photorealistic}. By breaking up the description into sentence chunks, we aim to study IIW's salient description style and also debias our results from description length. We evaluate $\sim$1k generated images across the varied input sentence chunks (over 240 random LocNar images) with a 3-way human ranking evaluation and CLIP similarity between the original and reconstructed image~\cite{radford2021learning}. 

The results in Tab.~\ref{table:reconstruction-results} indicate that IIW's detailed outputs consistently lead to better T2I reconstruction, with highest mean rank and CLIP similarity regardless of the length of input units. These results confirm that IIW descriptions capture the most visual content with the most detail, and that it is not strictly due to description length, but rather the saliency, comprehensiveness, and specificity in \textit{each} sentence that makes IIW impactful. As input text length is still a limitation in popular VLMs like CLIP, these results provide evidence that using only the first sentence of IIW descriptions can still be useful and performant. In Fig.~\ref{fig:reconstruction} we show examples of each model's description's resulting generated image and associated rank. Additional plots and examples are shared in Appendix~\ref{subsec:addtnt2i}.

\subsection{Compositional Reasoning with IIW}
\label{subsec:llmvqa}

We look to a second downstream evaluation to quantify the impact of our hyper-detailed image descriptions. Specifically, we use IIW generated descriptions to aid in vision-language compositional reasoning. Probing datasets ARO~\cite{yarom2023read}, SVO-Probes~\cite{hendricks2021probing}, and Winoground~\cite{thrush2022winoground} modify image captions to no longer match the paired image\footnote{SVO-Probes has a negative \textit{image} for each positive image-caption pair. The negative images also have captions, so we use those in our experiments.}: changing visual attributes or relationships, swapping verbs, or shuffling image captions such that they contain the same words but reflect different semantics. This is done to evaluate different types of vision-language reasoning, \eg, visual attribute understanding or verb understanding.

In this experiment we evaluate if IIW descriptions can be used to distinguish the real image caption from the incorrect negative caption in ARO, SVO-Probes, and Winoground datasets using an LLM-only setup. We prompt PaLM2-340B~\cite{anil2023palm} to select which of the caption options is true given the image description (see Appendix~\ref{subsec:appendixllmvqa} for exact input prompts). This essentially replaces the image in these datasets with a generated description; the amount the description is able to boost accuracy on these compositional reasoning tests should correlate to the description's comprehensiveness and specificity. We compare IIW fine-tuned models to two larger (7B) open source models: InstructBLIP-Vicuna-7B~\cite{dai2023instructblip} and LLaVA-V1.5-7B~\cite{liu2023visual} in Tab.~\ref{table:llm-vqa}, with additional models in  Appendix~\ref{subsec:appendixllmvqa}.

\setlength{\tabcolsep}{3pt}
\begin{table}[tb]
\centering
\begin{tabular}{lcccc}
\hline
Image Desc.& \multicolumn{2}{c}{ARO} & SVO- & Wino- \\
Model & VG-A & VG-R & Probes & ground\\
\hline
None & 56.50 & 59.94 & 50.71 & 49.88 \\
InstructBLIP-\textit{7B} 
& 83.99 & 62.73 & \textbf{89.35} & 65.25 \\
LLaVA-V1.5-\textit{7B}
& 84.80 & 63.71 & 87.89 & 63.38\\
IIW PaLI-3 \textit{5B} 
 & \textbf{90.37} & \textbf{66.19} & 88.66 & \textbf{69.38}\\
\hline
\end{tabular}
\caption{Vision-Language Compositional Reasoning Accuracy with Image Descriptions. We see if richer IIW descriptions can help distinguish the true matching image caption in ARO~\cite{yuksekgonul2023visionlanguage}, SVO-Probes~\cite{hendricks2021probing}, and Winoground datasets~\cite{thrush2022winoground}. COCO and Flickr30k Order subsets of ARO are not reported due to a very high language bias baseline of 98\%.}
\label{table:llm-vqa}
\vspace{-3mm}
\end{table}

Our first baseline is the no-image condition (None in the first row of Tab.~\ref{table:llm-vqa}), which simply asks an LLM which image caption is more likely. This serves an important language-bias baseline, and quantifies whether the vision-language compositional reasoning task really requires vision at all. Our results show that SVO-Probes and Winoground have the lowest language bias (baseline performs nearly at random). On the other hand, ARO visual genome attribution and relation subsets are not quite at random baseline; we also note that we do not include the Flickr30k nor COCO order ARO subsets, as the LLM can distinguish the true caption at 98\% accuracy without any image description.

When incorporating image descriptions, all models perform significantly better than the language-bias baseline. The IIW model results in the best task performance for ARO Visual Genome Attribution and Relation (VG-A, VG-R) and Winoground, with accuracy gains of nearly 34\%, 6\%, and 20\%, respectively. Moreover, we can further boost performance compared to the InstructBLIP and LLaVA image captions: we improve reasoning accuracy by about 6\%, 2\%, and 4\% compared to the best image description model-based baseline. This reflects the richness of IIW across different parts of speech and comprehensiveness, as more attributes and relationships are captured and can be used to reason about image content. For SVO-Probes, we find smaller differences, with IIW, InstructBLIP, and LLaVA models within $\sim$1 point of each other.
\section{IIW-Eval Benchmark Release}
\label{sec:data-release}

\setlength{\tabcolsep}{3pt}
\begin{table}[tb]
\centering
\begin{tabular}{lccccc}
\hline
IIW-Eval & IIW & \# &\multicolumn{3}{c}{Annotation Type} \\
Subset & Source & Images & Task-1 & Task-2 & SxS\\
\hline
IIW-400 & Human & \multirow{2}{*}{400} & 1,899 & 400 & 200 \\
 & Model &  & -- & 100 & -- \\
DCI & Human & 112  & -- & 112 & 112\\
DOCCI & Human & 100  & -- & 100 & 100 \\
LocNar & Model & 1000  & -- & 1000 & -- \\
XM3600 & Model & 1000  & -- & 1000 & --  \\
\hline
Total & & 2,612 & 1,899 & 2,712 & 412
\end{tabular}
\caption{IIW-Eval Data and Annotation Breakdown.}
\label{table:data-info}
\vspace{-2mm}
\end{table}

We release the \textbf{IIW-Eval} benchmark (Tab.~\ref{table:data-info}) of human- and model-annotated image descriptions, human SxS results on Human-Human and Model-Human pairs of descriptions. \textit{IIW-400} is a new eval set of 400 images randomly sampled from DOCCI-AAR~\cite{docci}. We re-annotate DCI and DOCCI test samples and enrich two existing datasets with new IIW descriptions: Localized Narratives (LocNar~\cite{PontTuset_eccv2020}) and CrossModal-3600 (XM3600~\cite{thapliyal2022crossmodal3600}). We provide LocNar and XM3600 annotations with significantly improved quality (see statistics in Appendix~\ref{sec:enriched}). The model generated descriptions may have hallucinations, information recall losses, or non-human like writing style artifacts. By releasing this subset along with human SxS judgements, we encourage the development of new metrics and evaluation systems to detect them in an automated, scalable manner. It also promotes fair comparison across methods in future work. 
The dataset is released under a \href{https://creativecommons.org/licenses/by/4.0/}{CC BY 4.0} license.
\section{Future Work} 

In future work, robust and effective automatic metrics are needed to evaluate the quality of detailed image descriptions. Next steps may include training model-based metrics or preference models (\ie, autoraters) with human preference data to learn a global quality metric. For additional analysis, we could further break down our current SxS metrics. For example, the human SxS hallucination metric could be broken down to capture fine-grained categories like how many hallucinations are with respect to color, size, or spatial location.

We are working to extend the ImageInWords framework to additional languages and geographically diverse images. In next steps, we note that images need to be sampled globally (across both geographic and cultural identity); this sampling must also be done across different image topics and categories, making equal coverage more complicated. We are currently working on adapting our proposed framework to accommodate locale specific annotators, which are required for cultural specificity. Our continued goal is to make the annotation guidelines holistic, reduce human effort and dependency in the annotation process, and help shift the narrative from captions to descriptions.

\section{Conclusion}
\label{sec:conclusion}

In this work, we proposed ImageInWords (IIW), a new framework for hyper-detailed image descriptions. Our annotation guidelines and seeded, sequential annotation process lead to human authored descriptions that are strongly preferred over both prior work's human annotations (+66\%) and prior work's fine-tuned models (+31\%). Images reconstructed with IIW generated descriptions were ranked 1st more often, regardless of how much of the image description was used, reflecting higher saliency earlier and better overall quality. Our compositional reasoning evaluation showed IIW generated descriptions to best contain fine-grained visual detail needed to decipher true from false visual attributes and semantics, with accuracy gains of up to 6\% over our most performant baselines. Our results collectively demonstrate the quality and utility of IIW image descriptions as state-of-the-art.
\clearpage
\section*{Limitations}
\label{sec:limitations}

Finally, we discuss the limitations of our annotation framework and evaluations. In our annotation framework, we define a seeded and sequential annotation process, with both aspects having potential limitations. The quality of the seeded data is of high importance as it will ultimately affect the rest of our human annotation pipeline. Additionally, even with the best possible seeds, they may limit the scope of what our crowd workers write by biasing them towards certain objects or phrases. We employed an active learning loop to iteratively improve the seed generation quality but significant room for improvement still remains.  In terms of limitations for the sequential augmentation used, unnecessary time may be spent by annotators if the first annotator output quality is low. By training the annotators through guidelines and feedback and monitoring the initially drafted descriptions, quality can be better ensured so that the framework is as efficient as possible.

With respect to the evaluation of our human annotated data and model generated outputs, we do only perform evaluations on hundreds of samples (as opposed to thousands or more). This is largely due to the cost and time associated with human SxS evaluations for this task, but we note that IIW is rated marginally and substantially better at a much higher rate, which would likely scale to more samples. Our work is also inherently limited by the lack of automated metrics available for long descriptions. We still report standard text similarity metrics in Appendix~\ref{sec:textsimmetrics} and complement them with human SxS, but in future we hope metrics are developed that address the current limitations, as automated metrics can be applied at scale. We note that metric limitations were also faced in prior work, with others opting to use LLM summaries or human SxS for evaluation purposes~\cite{urbanek2023picture,docci}. 

With respect to our trained IIW models, we also note that all results are reported from a single model/run for each evaluation included. In the future, rerunning models with different seeds or aggregating results over different model variants would be beneficial.

While we currently do not plan to open source our models or training set, we do release an evaluation set over images that can serve as a unified benchmark for IIW, recent, and future related work. We also open source the human SxS judgements and model enriched samples from Localized Narratives and XM3600. We acknowledge that the full annotation framework would take substantial time and effort to rerun from scratch; this is in part due to needing to reproduce the annotation UI and infrastructure for seeding. The framework itself is agnostic to which vision-language models are used for seeding of initial object or image captions, which we hope makes the setup more feasible to reproduce with any open source model of choice. This also becomes increasingly important as new and improved models will continue to be developed, and we’d like our framework to be able to incorporate newer models over time. The number of annotation rounds, annotation volume, and particular set of images can be adjusted to specific use-cases and budget and time constraints.

Lastly, our initial IIW dataset and resulting models are English-only. In the future, we plan to expand our work to have multilingual and multi-cultural coverage over images sampled globally. We also aim to curate images descriptions which are annotated by locale specific annotators to capture regional and cultural nuances, so that we do not strictly have descriptions with a western lens. 
\section*{Ethics Statement}
\label{sec:ethics}

Our model may have broader societal impact. It may contain unknown biases or stereotypes, or propagate inaccurate or otherwise distorted information. 
We used a combination of algorithmic methods, manual inspection, and other classifiers for identifying and removing Sensitive Personally Identifiable Information, pornographic, and violence depicting images. Specifically we checked for the presence of: 
(1) any address, email, or phone number; 
(2) images with high porn scores; 
(3) images labeled as portraying abuse; 
(4) text identified as having certain adult content references. 
Additionally, we asked human annotators to use an objective and respectful tone while composing the image descriptions. While we made all of these efforts, it is still possible the model may produce some undesirable results.

Additionally, image to text VLMs inherently can have negative impact if the generated image descriptions are inaccurate and/or contain hallucinations. However, our work specifically aims to cover all visual content as comprehensively and accurately as possible to improve data quality and the resulting fine-tuned models.

\bibliography{main}

\appendix

\section{Annotation Guidelines}
\label{sec:supplemental-guidelines}

We now present the full detailed annotation guidelines used for IIW annotations. Our guidelines state that image descriptions should be composed such that they \textit{paint a vivid mental picture} of an actual image in the mind of someone hearing the description and has their eyes closed. 
In order to reach this level of detail composed in an articulate manner, we compile an extensive set of annotation guidelines. We iterated over these guidelines with multiple pilot rounds.

The annotators are asked to {\it operate as if they are instructing a painter to paint with their words} and only include details that can be deduced from visual cues, erring on the side of higher precision. Unnecessary fragmentation of sentences should be avoided to compose writing in a flowy, coherent style, avoiding the use of \textit{filler phrases} like: ``\textit{In this image,}'' ``\textit{we can see,}'' ``\textit{there is a,}'' ``\textit{this is a picture of,}'' since they add no visual detail and come at a cost of verbosity.

\textit{Objects} form the lego-blocks of an image. Interactions and spatial arrangements among them help to form the context of the image. In complex multi-object images with dense settings, noting each and every object independently can become cumbersome and highly dependent on the effort the particular human annotator puts in. To define this better and expect a consistent behavior from the annotation outputs, we introduce the notion of \textit{salient objects}. Key objects without which the image would lose its context and meaning are considered \textit{salient}. This can include individual objects or combinations of them depending on the role they play in the image; consider the following 2 cases as examples:

\begin{itemize}
\item Three people in the blurry background of an image, with the scene set inside a coffee shop, who play no concrete role individually can be grouped as \textit{people in the background} instead of 3 individual \textit{people} object annotations.
\item Two people in the foreground and in-focus, engaged in a conversation in the same scene. The two individuals are likely the focus of the image and hence worth noting individually in detail as separate objects. This is likely what the photographer was attempting to capture.
\end{itemize}

While annotating each of these \textit{salient objects} in an image, the annotators should consider the following axes as reference (but not limit themselves to this list), paying special attention to features that make them unique or salient:

\begin{itemize}
\itemsep0em 
\item \textbf{\textit{Function}} Purpose of the component or the role it plays in the image
\item \textbf{\textit{Shape}} Specific geometric shape, organic, or abstract
\item \textbf{\textit{Size}} Large, small, or relative size to other objects
\item \textbf{\textit{Color}} Specific color with nuances like solid or variegated
\item \textbf{\textit{Design/Pattern}} Solid, flowers, or geometric
\item \textbf{\textit{Texture}} Smooth, rough, bumpy, shiny, or dull
\item \textbf{\textit{Material}} Wooden, metallic, glass, or plastic
\item \textbf{\textit{Condition}} Good, bad, old, new, damaged, or worn out
\item \textbf{\textit{Opacity}} Transparent, translucent, or opaque
\item \textbf{\textit{Orientation}} Upright, horizontal, inverted, or tilted
\item \textbf{\textit{Location}} Foreground, middle ground, or background
\item \textbf{\textit{Relationship to other components}} Interactions or relative spatial arrangement
\item \textbf{\textit{Text written on objects}} Where and how it's written, font and its attributes, single/multi-line, or multiple pieces of individual text
\end{itemize}

Humans typically associate a set of default features to objects. Consider the following examples:

\begin{itemize}
\item \textit{Car} by default is assumed to have 4 of each: tires, door, windows and 1 of each: trunk, hood, steering wheel, roof. Mentioning them separately might not be that useful as it adds no specific visual detail that we did not already know as the norm. Now, if the car is a \textit{coupe}, has a missing window, or contains a door painted with a different color than the overall color, \ie, making it a unique feature, then that would be worth mentioning in the description since it holds specific added visual value.
\item 
\textit{The Golden Gate Bridge} by default is orange. That being said, it does not hurt to include extra detail depending on the use-case. If the annotators do not recognize the bridge as a famous well known entity, then it would make sense to include the color and additional attributes.
\end{itemize}

When composing the overall image description, start with a newspaper style \textit{\textit{tldr}} sentence that paints a very clear high level picture. Describe the \textit{objects} in order of their \textit{saliency} while noting the description of individual objects and relationships in a coherent manner. Include the overall setting, background, style, and consider:

\begin{itemize}
\itemsep0em 
\item \textbf{\textit{Overall composition}} Arrangement of the elements in the image, focal point, balanced, or asymmetrical
\item \textbf{\textit{Lighting}} Natural or artificial, light source
\item \textbf{\textit{Color palette}} Colors or how they interact with each other
\item \textbf{\textit{Texture}} Smooth or rough, shiny or dull
\item \textbf{\textit{Depth of field}} Entire image or only a portion of it is in focus, what effect this has on the overall composition
\item \textbf{\textit{Subject matter}} Main subject of the image, other elements that are present, how they relate to the subject matter
\item \textbf{\textit{Mood or feeling}} Overall mood or feeling of the image
\end{itemize}

\begin{figure*}[tb]
\centering
\begin{subfigure}{\linewidth}
\includegraphics[width=\linewidth]{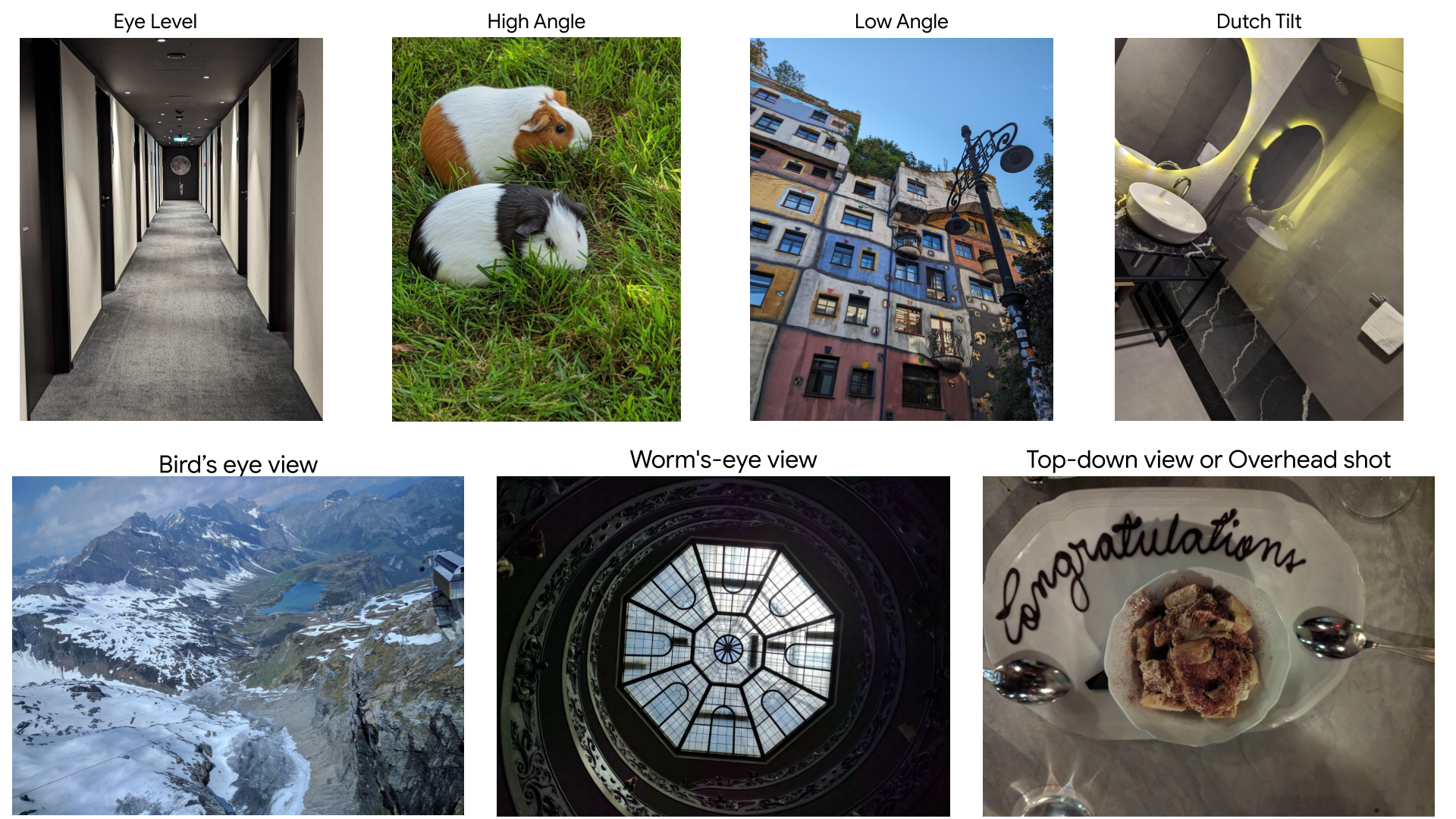}
\label{fig:ImageInWords_Figures-training-tasks}
\end{subfigure}
\caption{Camera Angles to Consider when Annotating Images. These are important to set a precedence on the level and kind of information to expect in the image description.}
\label{fig:camera_angles}
\end{figure*}

\textbf{\textit{Camera angle}} (\ie, the position of the camera in relation to the subject) is crucial, as this sets a precedence for what level and kind of information to expect. The choice of camera angle can have a significant impact on the mood and meaning of a photograph. Different camera angles can be used to create different effects and convey different messages, \eg, details about a close-up are different from those of a wide angle shot.
Examples of camera angles (see Figure~\ref{fig:camera_angles}):
\begin{itemize}
\item \textbf{Eye level:} The camera is positioned at the same level as the subject's eyes. This is the most natural and neutral camera angle.
\item \textbf{High angle:} The camera is positioned above the subject. This angle can make the subject appear smaller, weaker, or less important. 
\item \textbf{Low angle:} The camera is positioned below the subject, anywhere below the eye line, looking up. This angle can make the subject appear larger, stronger, or more important. Sometimes, it is even directly below the subject's feet.
\item \textbf{Ground level:} The camera is positioned at the ground level. This angle captures what is in the frame at ground level, that is, the feet, or maybe the character lying on the ground.
\item \textbf{Dutch tilt:} The camera is tilted on its axis. This angle can be used to create a sense of unease or disorientation.
\item \textbf{Bird's-eye view:} The camera is positioned directly above the subject. This angle can be used to show the subject's relationship to their surroundings.
\item \textbf{Worm's-eye view:} The camera is positioned directly below the subject. This angle can be used to create a sense of awe or wonder.
\item \textbf{Top-down view or Overhead shot:} The camera is above the subject and you're taking the photograph downwards from straight above, and not at any kind of angle. It is typically closer to the subject than a bird's eye view (see Figure~\ref{fig:camera_angles} for comparison).
\end{itemize}

Some other terms that are sometimes used to describe camera angles and depths:
\begin{itemize}
\item \textbf{Close-up:} A close-up is a photograph that is taken from a very small distance. Close-ups can be used to show details that would not be visible from a further distance.
\item \textbf{Medium shot:} A medium shot is a photograph that shows the subject from the waist up or from the knees up. Medium shots are often used to show the subject's body language and facial expressions.
\item \textbf{Long shot:} A long shot is a photograph that shows the subject from a distance. Long shots can be used to show the subject's relationship to their surroundings.
\item \textbf{Full shot:} A full shot is a photograph that shows the subject's entire body. Full shots are often used to show the subject's height and stature.
\item \textbf{Over-the-shoulder shot:} An over-the-shoulder shot is a photograph that is taken from behind one person's shoulder, showing the other person in the foreground. Over-the-shoulder shots are often used to create a sense of intimacy or connection between the two people.
\item \textbf{Point-of-view shot:} A point-of-view shot is a photograph that is taken from the perspective of the subject. Point-of-view shots can be used to create a sense of immersion in the scene.
\end{itemize}

When \textbf{\textit{text}} is present, include detail such as whether the text is in a single line or spread along multiple lines, if text is in multiple lines whether there is mutual alignment, the features of the \textit{font} such as size, style, color, and orientation (\eg, vertical, horizontal, arched), \textit{casing} (\eg, lower, upper, mixed), and attributes like italics, underlined, bold, written in quotes, clearly visible or blurred. Describe the words if they are written.

If text is written in multiple lines, we should:

\begin{itemize}
\item Quote them as individual units that exist on the same line
\item Mention its mutual alignment using references like vertically stacked, aligned to the left, \etc
\end{itemize}

\begin{figure*}[tb]
\centering
\begin{subfigure}{\linewidth}
\includegraphics[width=\linewidth]{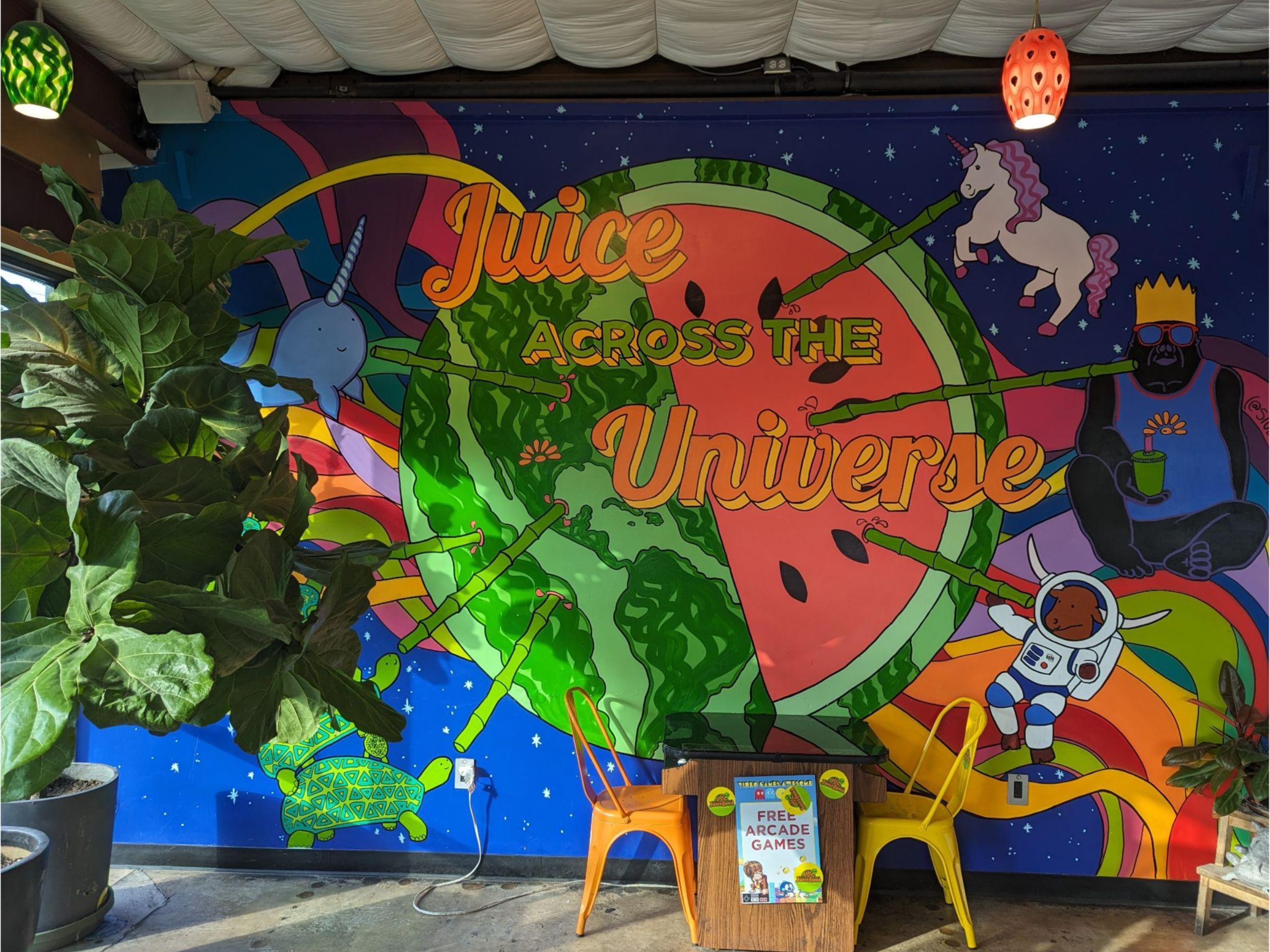}
\end{subfigure}
\caption{An Example where Quoting \textit{Text} in a Detailed Manner can Enable Precise Reconstruction. The word-casing and alignment attributes of the multi-line phrase (``Juice,'' ``ACROSS THE,'' ``Universe'') has words ``Juice'' and ``Universe'' as capitalized while the phrase ``ACROSS THE'' is all upper-cased and all components are aligned along a diagonal. Information on the font color, type, shadow effect should be included. For the phrase (``FREE,'' ``ARCADE,'' ``GAMES'') all words are upper-cased, vertically stacked, and centrally aligned.}
\label{fig:supplemental-text-annotation}
\end{figure*}

For example, in Figure~\ref{fig:supplemental-text-annotation}, the phrase (``Juice,'' ``ACROSS THE,'' ``Universe'') has words ``Juice'' and ``Universe'' as capitalized while the phrase ``ACROSS THE'' is all uppercase, and components are aligned along a diagonal. Information on the font color, type, and shadow effect should be included. As another example from the same image, the phrase (``FREE,'' ``ARCADE,'' ``GAMES'') are all upper-cased, vertically stacked and centrally aligned.

If you have a good idea of the font family and are confident, that would be valuable to note.

When \textbf{\textit{people}} are present, special notes should be kept in mind to mitigate different types of bias. The tone should be \textit{respectful} to the subject and not make assumptions or try to guess their gender, identity, ancestry, where they are from, sexuality, religion, \etc. We emphasize that the descriptions should be noted in objective, neutral and fair language for related attributes and focus solely on the visual aspects. Consider the following axes with respect to attributes here:
\begin{itemize}
\itemsep0em 
\item How much of their body is visible
\item Whether the face is fully visible
\item Whether they are facing the camera or looking somewhere else
\item Where and what they are looking at	
\item What the person is doing (standing, posing, sitting, running, playing a sport)
\item What they are wearing. For each piece, note the clothing item name (dress, pants, short, gloves, shoes), color, pattern (plain, striped), length (if applicable)
\item What they are carrying, details about that object (bag, purse, camera)
\item Whether they are using any assistance device (wheelchair, cane)
\item Whether they have any unique features like marks, tattoos, scars on their body that are visible. If applicable, note the respective positions on their body where each is present
\item For professions with known gender biases like ``nurse,'' ``doctor,'' or ``construction worker,'' explicitly include the gender (if clearly deducible) and do not operate under the assumption that one gender is more common in that profession.
\end{itemize}

For any \textbf{\textit{apparel}}, the descriptions should focus on overall style, unique details, silhouette of the garment, how it fits, fabric, color, shades, and tone of the garment. If the branding is visually visible, it should be included while attributes like size should be skipped unless visually verifiable.

Where applicable use \textbf{\textit{locale specific names}} of objects like \textit{clothing} (\eg, sherwani, kurta, kimono, saree), \textit{food} (\eg, shawarma, dosa, paneer tikka) \etc. The aim is to capture the locale specific vocabulary so the downstream models can pick them up instead of using generic abstract terms.

For \textbf{\textit{art pieces}}, include art styles, time periods, mediums, moods, viewpoints, subject matters, cultures as much as possible from the visual cues.
\section{Dataset Collection}
\label{sec:supplemental-dataset}

\begin{figure*}[t!]
\centering
\begin{subfigure}{\linewidth}
\includegraphics[width=\linewidth, keepaspectratio]{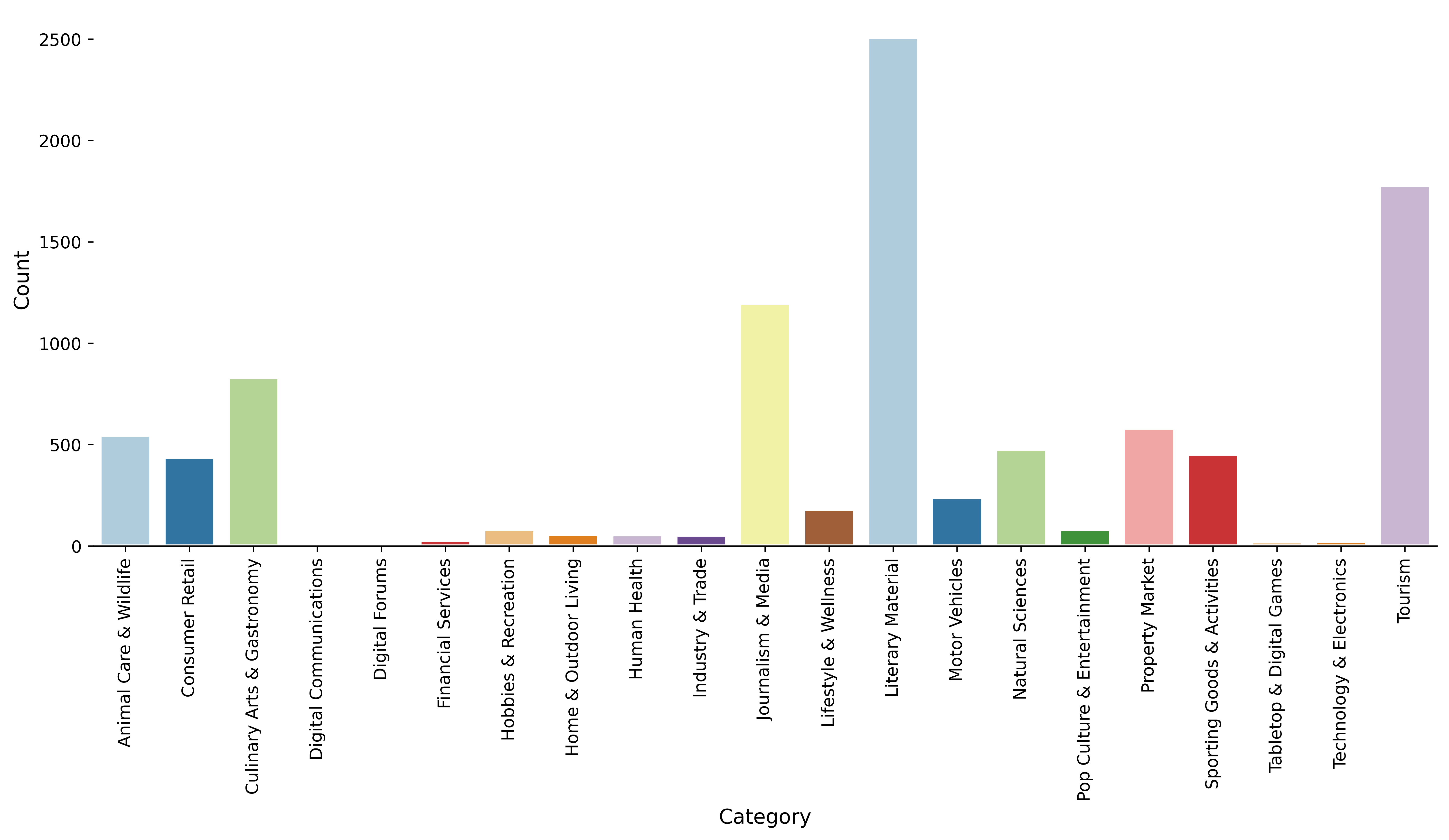}
\caption{IIW-Train Set Image Category Distribution}
\end{subfigure}
\begin{subfigure}{\linewidth}
\includegraphics[width=\linewidth, keepaspectratio]{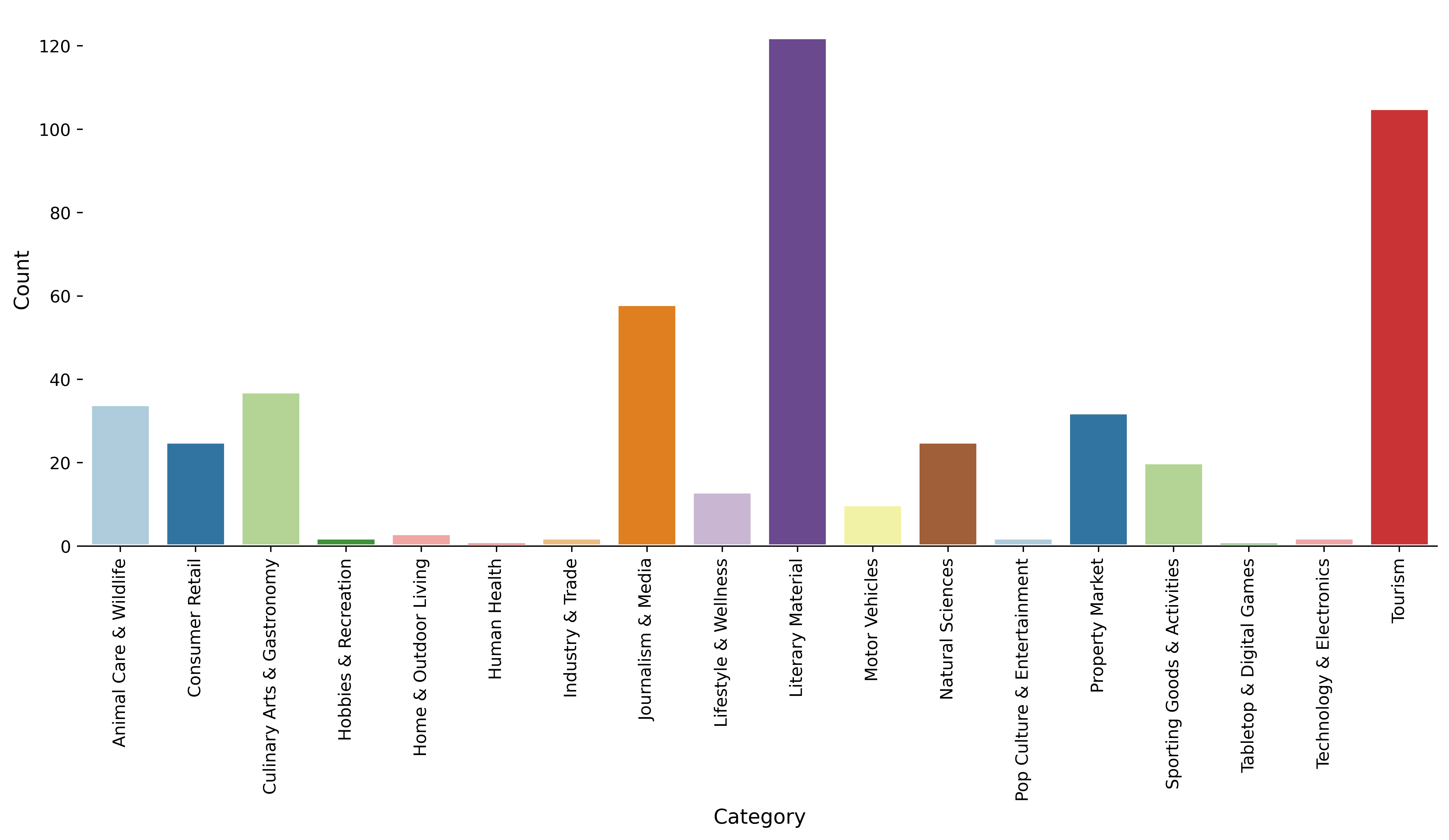}
\caption{IIW-Eval Set Image Category Distribution}
\end{subfigure}
\caption{Image Category Distribution for the IIW Dataset's Train and Eval Splits.}
\label{fig:supplemental-dataset-categories}
\end{figure*}

The dataset was sampled to cover a wide range of content. We use an internal image classification system to report the top image categories present across the splits in Figure~\ref{fig:supplemental-dataset-categories}. Getting a more balanced mix remains active work on our part and would be updated in future work.

\subsection{Human Annotation Worker Pool}
\label{subsec:annpool}
We employed and worked with a fixed human annotator pool comprising of 20+ annotators with mixed backgrounds in creative writing, art, history, photography and related relevant domain subjects to utilize critical domain expertise and perspectives. The pool is based in multiple countries, with a US majority currently. In the future, we plan to intentionally increase diversity in our annotator pool to ensure more locale-specific vocabulary in our image descriptions.
The annotators were compensated appropriately taking their skill-set, qualifications, location and the complexity of the task into account.
The pool was trained for the annotation task over a period of month to achieve a sense of consistency on the annotation guidelines as well as the downstream tasks to be covered by the data being collected. The annotators were also communicated clearly on the downstream tasks and data use cases to get a sense of the importance and quality bar needed for this foundation work.
For text-to-image generation rankings, we employed an internal group of six people to rank the images generated by different model-generated image descriptions (\ie, we did not hire crowd workers). People participating are domain experts, familiar with text-to-image generation technology. 

\subsection{Human Annotation Challenges}
Despite the very detailed annotation guidelines we provided to the annotators, there were several challenges during the human annotation process. First, we still found individual instances of random quality or judgment lapses. To circumvent this, we designed our framework to be sequential (\ie, more than one annotator works on each sample). We also found different challenges with respect to each image. For instance, art images require more domain specific expertise to describe an image with appropriate vocabulary. At the start of our annotation process, we observed that annotators had a tendency to use filler words and prefixes such as ``\textit{This is a},'' ``\textit{There is a},'' or ``\textit{This photo was taken with},'' and we provided feedback asking they do not include such phrases.

Another challenge during the annotation process was to encourage annotators to focus on the big picture and write a TLDR first. We also observed some tendency to use slightly subjective language while describing the images, \eg using adjectives that are not explicitly supported by the visual cues. By providing feedback directly to the annotators, pointing to specific samples, and emphasizing that certain language styles do not align with the writing style we were aiming for, we were able to considerably increase the annotation quality and get the desired type of image descriptions from the annotation process.

\subsection{Annotation Methodology}

\smallskip
\noindent\textbf{Seeded Annotation} Considerations to keep in mind:

\begin{enumerate}
\item \textbf{Quality of the seeding data} is critical. It is counter productive if it's noisy as the human annotators will take longer to comb signal from the noise than to come up with the information themselves. We recommend to restrict the use of seeding signal to only high precision models.
\item Risk of \textbf{biasing} the outputs as the human annotators may take the easy route of relying on the seed signal more heavily than intended. We suggest to note this point explicitly in the annotation guidelines and spot check the annotations for quality control. Additionally, running annotations with no seeding and comparing the outputs can be helpful to judge the bias being induced.
\end{enumerate}

\smallskip
\noindent\textbf{Sequential Augmentation} Considerations to keep in mind:
\begin{enumerate}
\item Heavy reliance on the quality of the base dense description from the first annotator. If the quality is not good, the annotator in the next round will spend considerable time fixing the input. There are 2 mitigating steps:
\begin{enumerate}
\item Monitor this at the beginning of the annotation project when the annotators are still new to the task using metrics like edit-distance and provide explicit feedback to the annotators as needed.
\item Annotators in each round have the option to start from scratch if they deem the quality from the previous round to be considerably low. Use this as feedback for the annotator from the previous round by presenting them the edited output to learn from.
\end{enumerate}
\end{enumerate}

\begin{figure*}[t]
\centering
\begin{subfigure}{\linewidth}
\includegraphics[width=\linewidth]{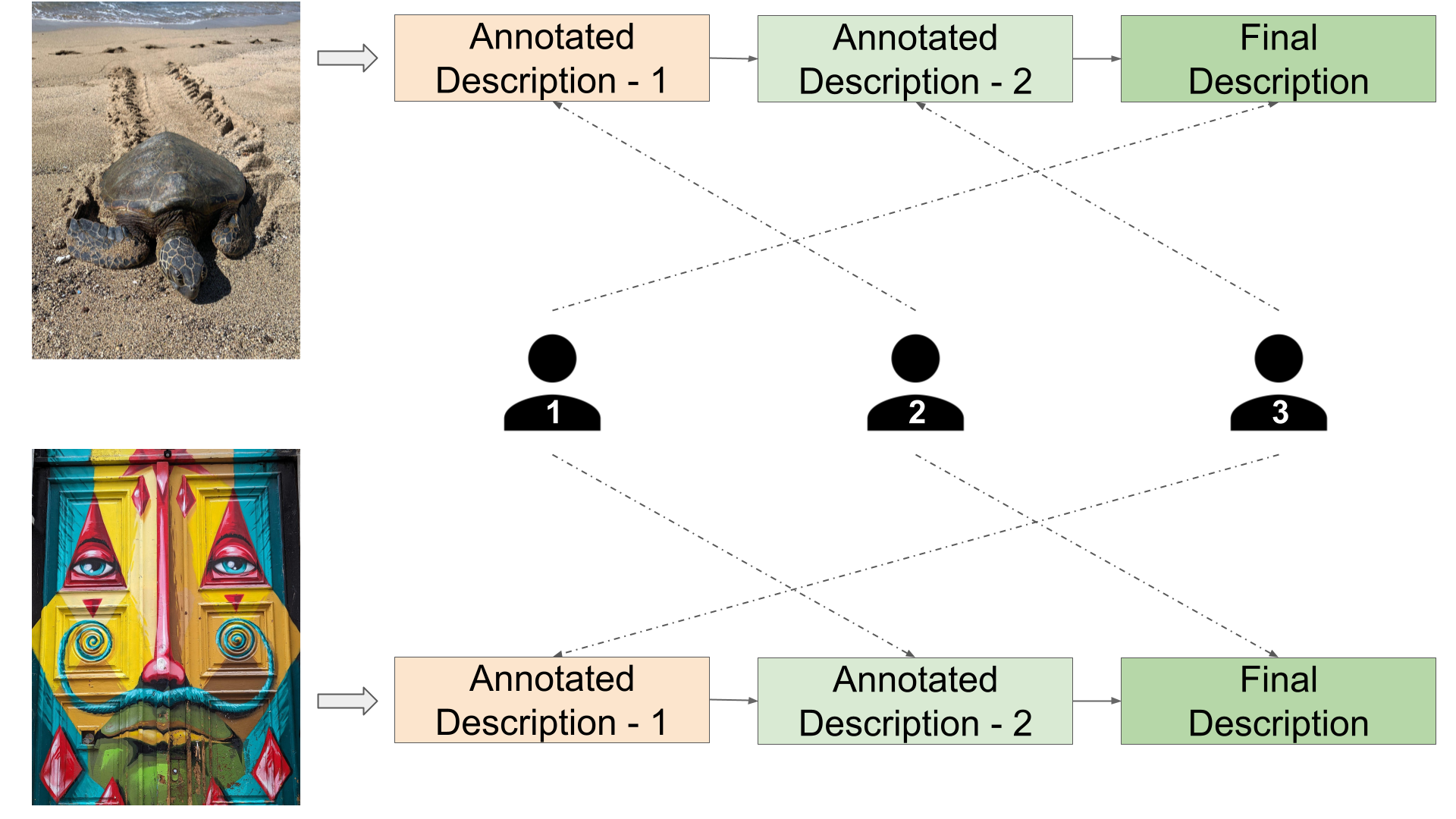}
\end{subfigure}
\caption{Human-in-the-Loop Learning. Over time with a constant annotator pool, each annotator gets an opportunity to read and learn from others' perspective via an \textit{implicit feedback loop}. This has shown to improve individual annotator quality as shown in the main paper.
}
\label{fig:supplemental-human-in-the-loop}
\end{figure*}

\smallskip
\noindent\textbf{Human-in-the-Loop Learning}
Our annotation framework implicitly unlocks a feedback loop for the annotators due to the sequential augmentation process discussed above. Each annotator gets an opportunity to read and learn from each other's perspective which in turn improves their individual quality. As an example from Figure~\ref{fig:supplemental-human-in-the-loop}, we demonstrate how Annotator-1 get an opportunity to learn from Annotator-3 for the first image and Annotator-2 gets an opportunity to learn from Annotator-1 in the second image.

\smallskip
\noindent\textbf{Model-in-the-Loop Annotation}
We employ an active learning loop for the VLMs where after some initial annotation data is available, a model version \textit{M$_{1}$} can be trained over the base VLM to improve the seed description quality. As more data gets annotated, \textit{M$_{1}$} can be updated to \textit{M$_{2}$, M$_{3}$, ..., M$_{n}$} to reduce the human effort needed.

Advantages:
\begin{enumerate}
\item Reduces the dependency on the \textit{human} both in terms of number of annotation rounds and time.
\item Provides a way to evaluate current model quality by monitoring the time, volume and patterns of augmentations during the human annotation stage.
\end{enumerate}

Some considerations to keep in mind:
\begin{enumerate}
\item As discussed above, the effectiveness relies very heavily on the capability of the model, \ie, having high comprehensiveness and low hallucinations.
\end{enumerate}

\subsection{Annotation Framework}
\label{subsec:uiexamples}
We now discuss the annotation framework with concrete examples and UI illustrations:

\begin{figure*}[t!]
\centering
\begin{subfigure}{\linewidth}
\includegraphics[width=\linewidth]{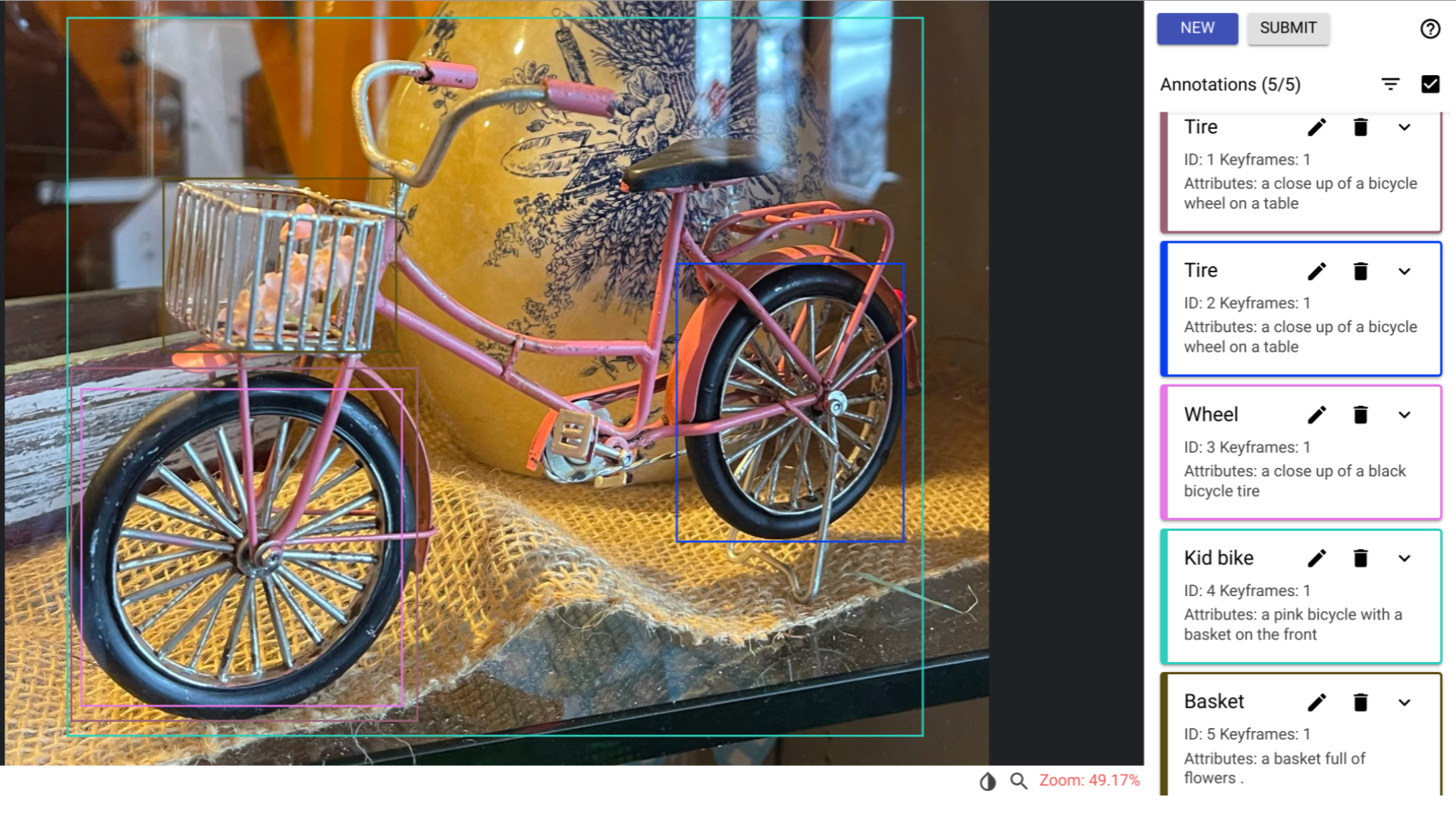}
\end{subfigure}
\caption{IIW Annotation UI for Task-1 with VLM seeds. We illustrate the seed object-detection objects and VLM generated object-level captions with object cropped image bytes as input.}
\label{fig:suplemental-data-augmentation-example-task-1-seeded}
\end{figure*}

\begin{figure*}[t!]
\centering
\begin{subfigure}{\linewidth}
\includegraphics[width=\linewidth]{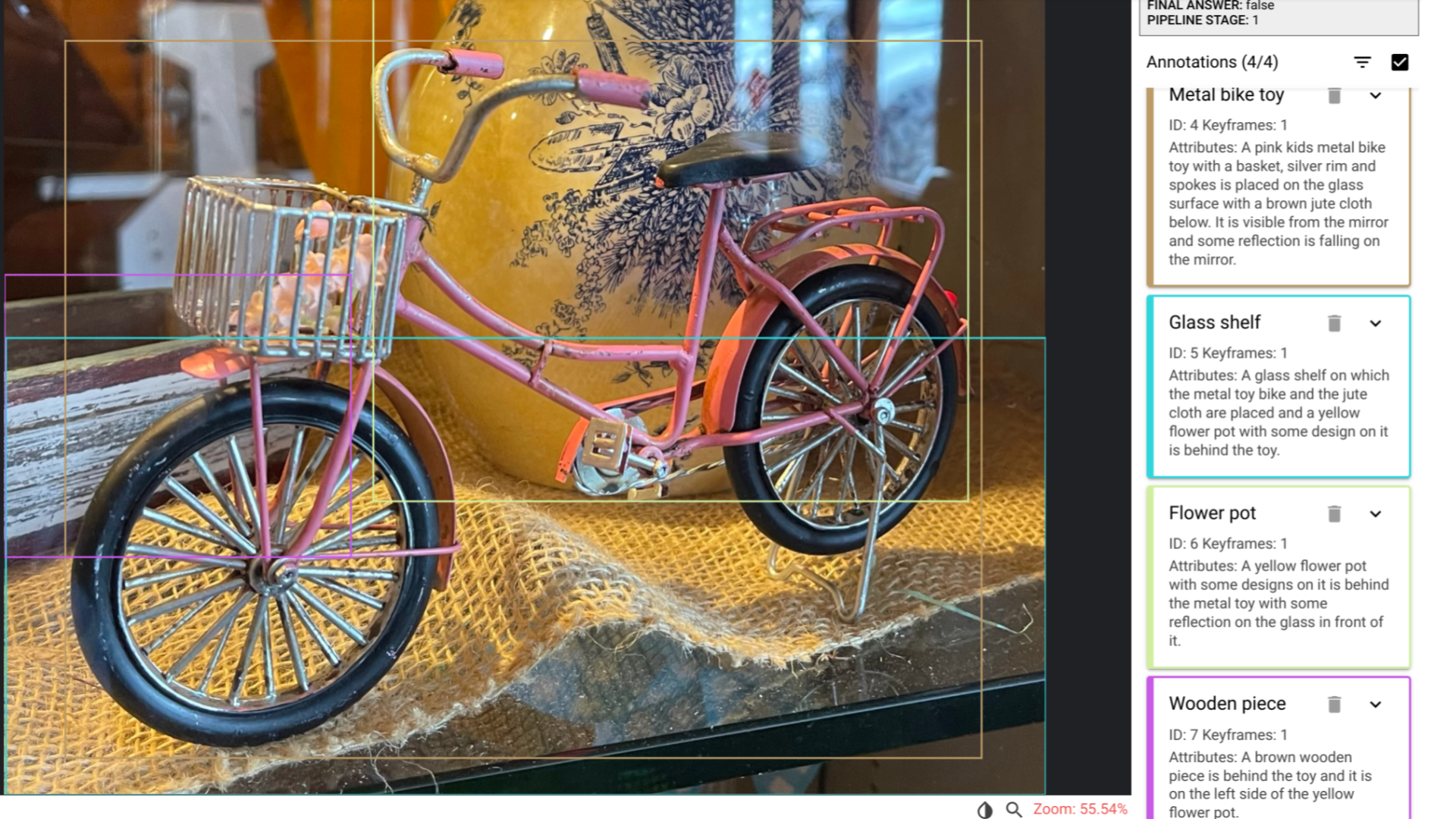}
\end{subfigure}
\caption{IIW Annotation UI for Task-1 after human augmentation. We illustrate the human augmented salient objects and their human-authored descriptions. The annotations are built on seed information from Figure~\ref{fig:suplemental-data-augmentation-example-task-1-seeded}. This example demonstrates how humans can alter the seed annotations based on the annotation guidelines, which can include merging, deleting, editing and adding new salient objects and then describing each.}
\label{fig:suplemental-data-augmentation-example-task-1-annotated}
\end{figure*}

\smallskip
\noindent\textbf{Annotation Task-1: Fine Grained Objects and Attributes}
In Task-1, the human annotators are presented with seed annotations for the objects from an Object-Detection (OD) model and VLM generated seed captions for each object (see Figure~\ref{fig:suplemental-data-augmentation-example-task-1-seeded}). The annotators can then annotate to note the salient objects and their corresponding description (see Figure~\ref{fig:suplemental-data-augmentation-example-task-1-annotated}).  

Annotators can make the following augmentations to annotate salient objects:

\begin{itemize}
\item  \textbf{Edit} make adjustments to the label and/or bounding box. This can include:
\begin{itemize}
\item Making the labels more specific, e.g \textit{Animal} to \textit{German Shepherd}
\item Enlarging or tightening the bounds of the bounding box by expanding or contracting the seed box.
\end{itemize}
\item  \textbf{Remove} any invalid pre-populated objects or considerably invalid bounding boxes.
\item  \textbf{Add} any missing salient object by drawing out a tight bounding box and adding an appropriate fine-grained label to it.
\item  \textbf{Merge} if object(s) are fragmented and/or pre-populated as two or more objects, the annotators can remove the individual objects and create a new single object.
\begin{itemize}
\item Closely placed objects of the same/similar label/type which individually hold low value but can be described as a collection to hold a higher context value should be combined, \eg, five identical cups in an image lined up next to each other do not need to be tagged as separate objects. If there are attributes that separate one or more of them from the others, we expect the annotators to split them in groups and proceed accordingly.
\item Sub-components of a larger object should not be explicitly tagged unless there is something unique and/or worth mentioning about them. Think \textit{does missing this detail create a different mental picture than the actual image?}, \eg, doors, windows, or tires of a \textit{Car} can be omitted unless there is something unique about them, as they are standard expectations from a \textit{Car} object.
\end{itemize}
\end{itemize}

For each (label, bounding box) pair, we ask the annotators to generate a detailed description focused on the object in the context of the image considering the several axes as reference (see Appendix~\ref{sec:supplemental-guidelines}). 

\begin{figure*}[t!]
\centering
\begin{subfigure}{\linewidth}
\includegraphics[width=\linewidth]{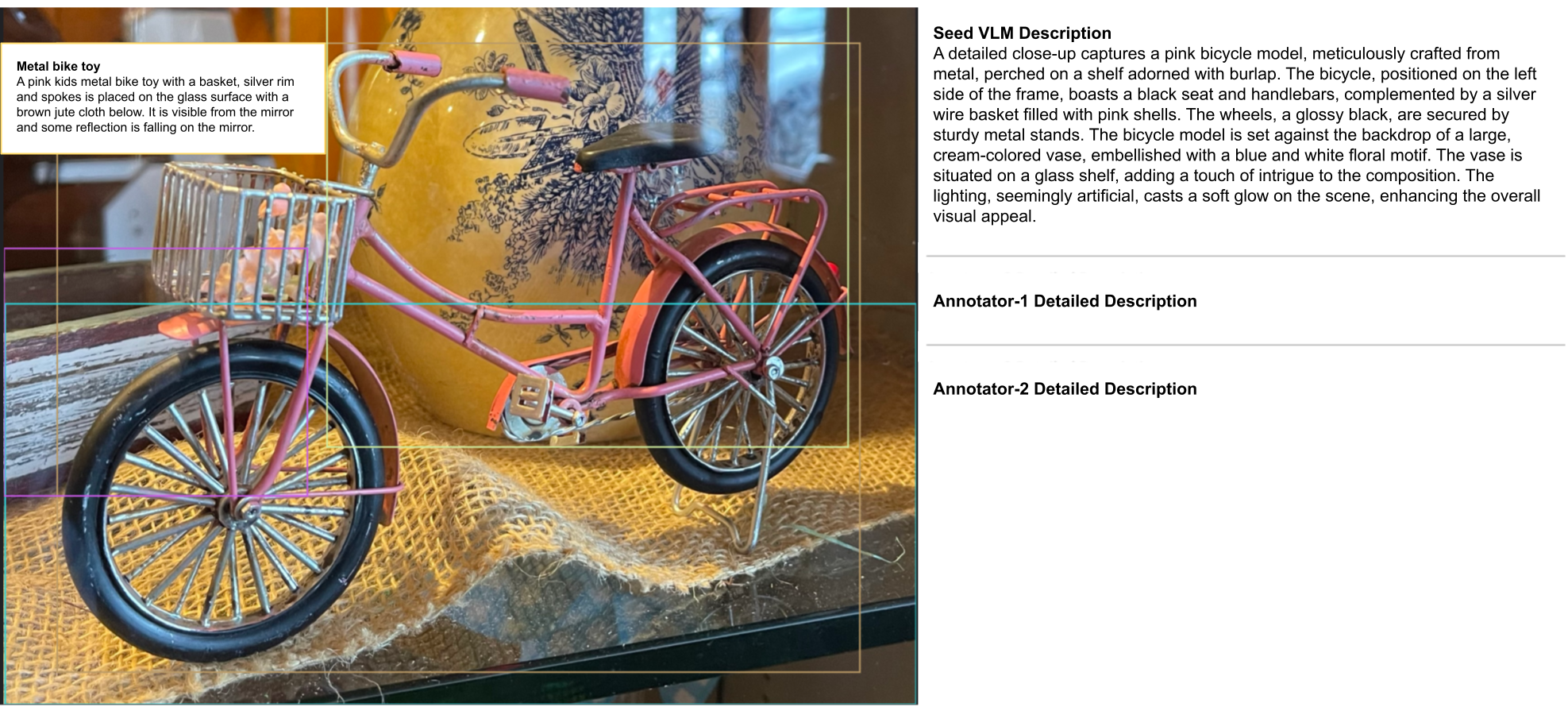}
\end{subfigure}
\caption{IIW Annotation UI for Task-2 with seed VLM description. This VLM has been fine-tuned in an active learning mode as data was collected iteratively. The seed caption from the same VLM (PaLI-5B) without the IIW fine-tuning is \textit{``a pink bicycle with a basket of flowers on it.''} The seed annotation is then refined and augmented by human annotators, see Figure~\ref{fig:supplemental-data-augmentation-example-task-2-annotated} for the final resulting description.}
\label{fig:supplemental-data-augmentation-example-task-2-seed}
\end{figure*}

\begin{figure*}[t!]
\centering
\begin{subfigure}{\linewidth}
\includegraphics[width=\linewidth]{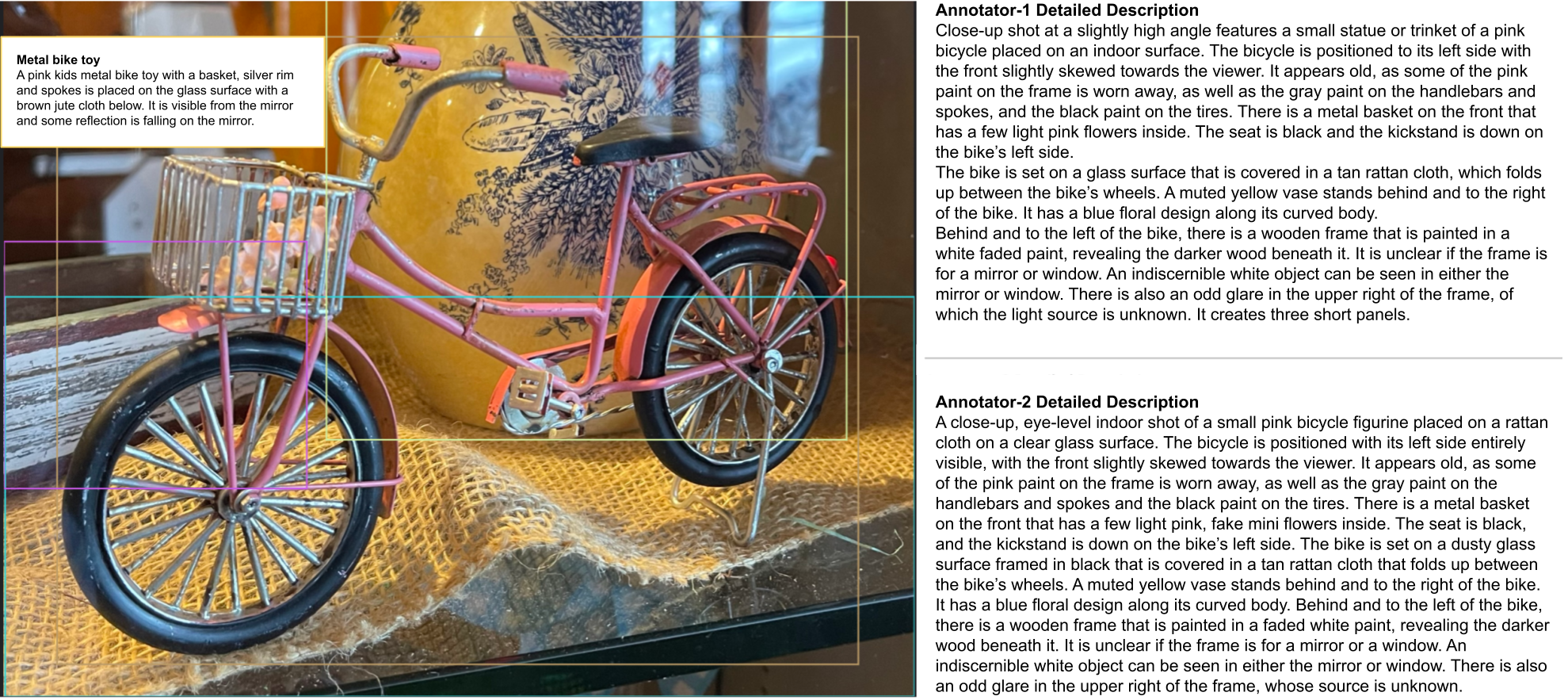}
\end{subfigure}
\caption{IIW Final Annotation UI for Task-2. We illustrate the human annotations available from Task-1 as the human annotators hover over the salient objects in the image. The annotators can additionally switch between hiding all salient objects to view the image properly. Task-2 annotations start with the seed caption from the VLM and is then refined by human annotators in sequential rounds, building on top of the previous round's output.}
\label{fig:supplemental-data-augmentation-example-task-2-annotated}
\end{figure*}

\smallskip
\noindent\textbf{Annotation Task-2: Overall Image Description}
In Task-2, human annotators are presented with the annotations from Task-1 and a seeded VLM description (see Figure~\ref{fig:supplemental-data-augmentation-example-task-2-seed}) which is then refined by human annotators in sequential rounds to produce the final hyper-detailed description (see Figure~\ref{fig:supplemental-data-augmentation-example-task-2-annotated}).

\section{IIW Fine-Tuning Tasks}
\label{sec:supplemental-tasks}

\begin{figure*}[tb]
\centering
\begin{subfigure}{\linewidth}
\includegraphics[width=\linewidth]{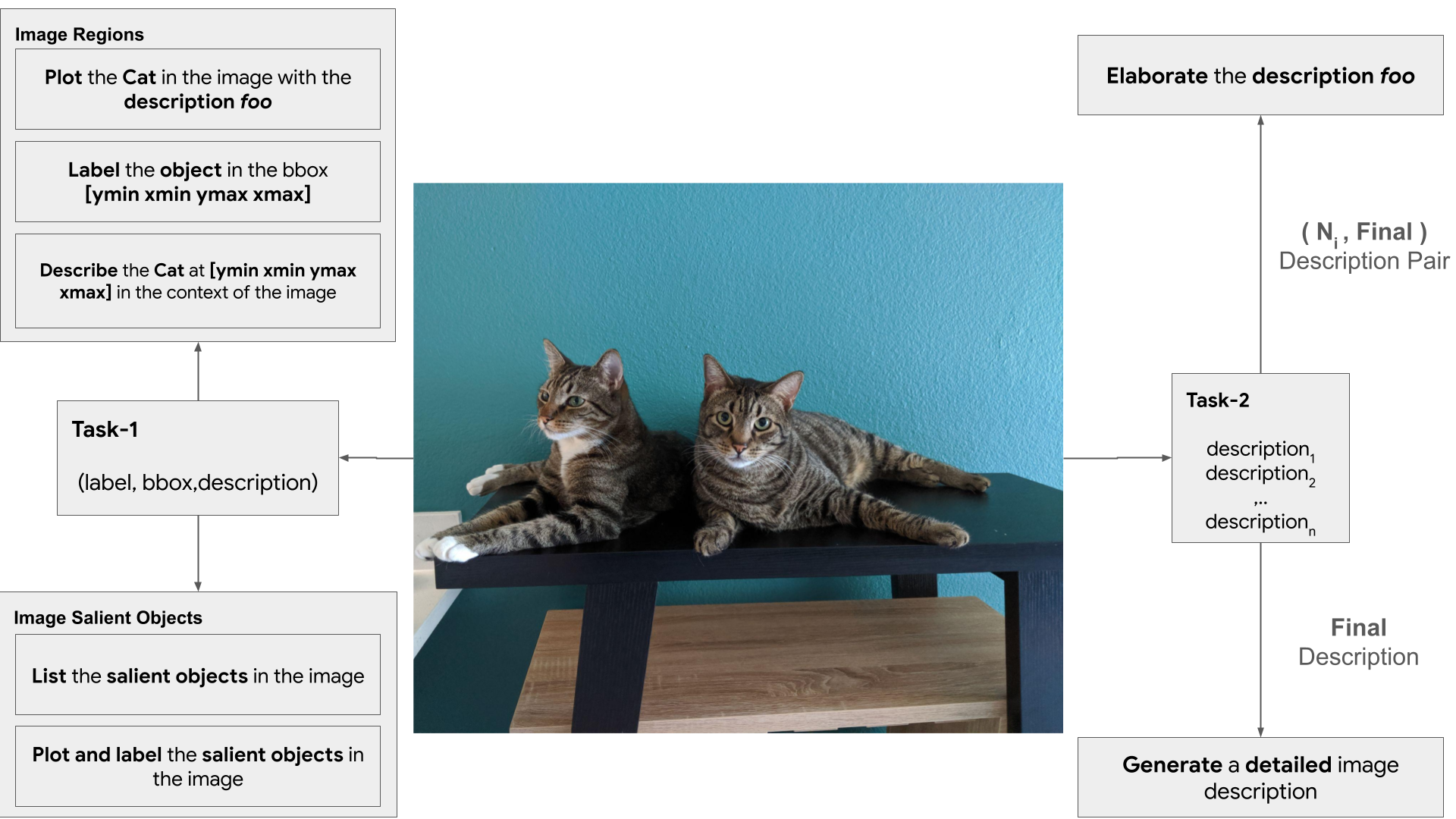}
\end{subfigure}
\caption{IIW based VLM Fine-tuning Tasks. We show tasks based on data collected from Task-1 and Task-2 per the IIW annotation framework. Different tasks enable the fine-tuning to focus on the image at (object, attribute), (image, objects) or (image, hyper-detailed description) levels.}
\label{fig:training_tasks}
\end{figure*}

We define seven tasks with the IIW Task-1 and Task-2 annotations to fine-tune two IIW based VLM model variants of PaLI-3 5B~\cite{chen2023pali3}. Our models include \textit{IIW Combined}, trained on a mixture of all seven tasks and \textit{IIW-Task-2} based aka \textit{IIW Model}, which is only trained on the final most detailed image description output. The seven tasks can be grouped into three categories: image region, salient objects, and detailed description based tasks, see Figure~\ref{fig:training_tasks} for illustration.

As we later discuss, we generally find the IIW (Task 2 only) Model to be preferred over the IIW Combined variant, but include details on the additional training tasks and resulting ablations here for completeness. All results in the main paper use the IIW Model.

\subsection{Image Region Tasks}
Using one object at a time from the list of \textit{(label, bounding box, description)} Task 1 annotations, we perform three region-based tasks. We use normalized bounding boxes in [\textit{ymin, xmin, ymax, xmax}] format as in Pix2Seq~\cite{chen2022pix2seq}. Our first task is description-label grounding. In multi-object dense images, a label in itself is not enough to uniquely identify an object. Thus, we create a grounding task with \textit{(image, label, description)} inputs that are tasked to predict the corresponding normalized bounding box coordinates.

Our second image region task is label prediction, in which we predict an open vocab label for the object with input \textit{(image, bounding box)}. Lastly, we perform object description generation, which produces descriptions for each object in the image given \textit{(image, bounding box, label)}.

\subsection{Salient Objects Tasks}
Our next category of fine-tuning tasks concerns the salient objects in an image. We target the aggregated list of \textit{(label, bounding box)} object features per image from Task 1. Our first task is label generation, in which given an image, we aim to generate a text list of the salient object labels. The object labels are sorted alphabetically for consistency, but in future work ordering by saliency would be useful. Our second object-level task is \textit{grounded} label generation. The task is to generate the list of \textit{(label, bounding box)} pairs per object in the image; we similarly sort the list alphabetically with respect to label name.

\subsection{Detailed Description Tasks}
Finally, our last fine-tuning tasks relate to the sequentially annotated descriptions from Task 2. We perform description elaboration in addition to direct description generation. Given the image and description from the $N^{th}$ sentence, description elaboration trains the model to elaborate the current description to the final description. We also create synthetically corrupted versions of the final description to serve as additional training samples. Specifically, we randomly drop \textit{X}\% of sentences. Sentences are dropped starting from the last sentence so that the structure of the overall text piece is maintained (as opposed to random sentence removal). For final description generation, given the image, a VLM learns to generate the final most hyper-detailed description available from the entire annotation framework. This final task (and not description elaboration), is the only task used to train the IIW model (whereas all are used for the IIW Combined ablation).

\section{Experiments}
\label{sec:supplemental-experiments}

\subsection{Seeded Annotation SxS}

\begin{table}[t!]
\centering
\begin{tabular}{l|ccccc}
\hline
\multirow{3}{*}{Metric} & \multicolumn{5}{c}{IIW-400}\\
\cline{2-6}
& \multicolumn{2}{c|}{Unseeded}&&\multicolumn{2}{|c}{Seeded}\\
\cline{2-6}
& \text{\tiny\faPlus\faPlus} & \text{\tiny\faPlus} & - & \text{\tiny\faPlus}&  \text{\tiny\faPlus\faPlus} \\
\hline
Comprehensiveness & 6 & 8 & 18 & \textbf{45} & 23 \\
Specificity & 10 & 6 & 20 & \textbf{39} & 25 \\
Hallucinations & 4 & 16 & \textbf{51} & 23 & 6 \\
TLDR & 4 & 27 & 10 & \textbf{43} & 16 \\
Human-Likeness & 10 & 12 & 31 & \textbf{33} & 14 \\
\hline
\end{tabular}
\caption{Human SxS to Evaluate Gains from Seeding the Annotation in the IIW Annotation Framework. We report rounded percentages comparing 50 IIW-400 samples annotated by the IIW framework  with and without machine-generated seeding on Comprehensiveness, Specificity, Hallucinations, TLDR quality, and Human-Likeness.}
\label{table:sxs-seed}
\end{table}

We additionally run a human SxS evaluation to compare the effects of seeding in the IIW annotation framework. In Table~\ref{table:sxs-seed}, we compare descriptions written without and with VLM seeding on a subset of IIW-400 (50 samples). There is a trend across all metrics that seeding improves description quality, as seen with marginal or substantial gains across comprehensiveness (+54\%), specificity (+48\%), TLDR quality (+28\%), and human-likeness (+25\%). The hallucinations metric is primarily neutral with a slight preference to seeded descriptions (+9\%). This is somewhat expected, and affirms that despite model-generated outputs having a potential risk for hallucinations, the humans are able to correct and improve on them. Thus, the SxS confirms seeding is advantageous to the IIW annotation framework.

\subsection{IIW Human versus IIW Model SxS}

In Table~\ref{tab:human-vs-model-iiw}, we perform a SxS evaluation on a subset of IIW-400 (on 100 samples). This compares data from the human authored IIW annotation framework to descriptions generated by the IIW fine-tuned model. Across all metrics there is an extremely high preference to the human annotated data, with significant and marginal gains: comprehensiveness (+78\%), specificity (+91\%), fewer hallucinations (+31\%), TLDR quality (+58\%), human-likeness (+52\%).  This confirms the quality of data produced by the IIW human-in-the-loop annotation framework, and demonstrates the need for more modeling efforts to bridge the gap between the IIW human authored versus model generated description quality. For example, larger capacity models may be needed.

\setlength{\tabcolsep}{7pt}
\begin{table*}[tb]
\centering
\begin{tabular}{l|ccccc}
\hline
\multirow{3}{*}{Metric} &\multicolumn{5}{c}{IIW-400}\\
\cline{2-6}
\cline{2-6}
&\multicolumn{2}{c|}{IIW-Human}& & \multicolumn{2}{|c}{IIW-Model}\\
\cline{2-6}
& \text{\tiny\faPlus\faPlus} & \text{\tiny\faPlus} & - & \text{\tiny\faPlus} & \text{\tiny\faPlus\faPlus}\\
\hline
Comprehensiveness & 40 & \textbf{43} & 12 & 4 & 1\\
Specificity & \textbf{79} & 14 & 5 & 2 & 0\\
Hallucinations & 6 & \textbf{46} & 33 & 17 & 4\\
TLDR & 29 & \textbf{43} & 14 & 10 & 4\\
Human-Like & 27 & 32 & \textbf{34} & 6 & 1\\
\hline
\end{tabular}
\caption{Human SxS to Evaluate IIW Fine-tuned PaLI-3 5B Model Predictions when compared to IIW Human-Authored Data on IIW-400 using 100 samples.}
\label{tab:human-vs-model-iiw}
\end{table*}

\subsection{Automatic Readability Measurements}
\label{subsec:readability-eval}

\setlength{\tabcolsep}{5pt}
\begin{table*}[tb]
\centering
\begin{tabular}{l|cccc|cccc}
\hline
\multirow{2}{*}{Dataset} & \multicolumn{4}{c|}{\textcolor{red}{Human} Authored}& \multicolumn{4}{c}{\textcolor{blue}{Model} Generated}\\
\cline{2-9}
& ARI$\uparrow$ & FK$\uparrow$ & GF$\uparrow$ & SMOG$\uparrow$& ARI$\uparrow$ & FK$\uparrow$ & GF$\uparrow$ & SMOG$\uparrow$\\
\hline
DCI & 5.8 & 5.7 & 8.1 & 8.1 & 2.9 & 3.7 & 6.2 & 6.9 \\
DOCCI & 7.5 & 7.1 & 9.5 & 8.7 & 6.4 & 6.6 & 8.7 & 8.2\\
IIW & \textbf{10.4} & \textbf{9.5} & \textbf{11.8} & \textbf{11.5} & \textbf{9.3} & \textbf{9.0} & \textbf{11.3} & \textbf{11.7}\\
\hline
\end{tabular}
\caption{Readability Metrics on Human and Model Annotated Data. We include ARI~\cite{enwiki:1145735758}, Flesch Kincaid (FK)~\cite{enwiki:1192056958}, Gunning Fog (GF)~\cite{enwiki:1181089308}, and SMOG~\cite{enwiki:1192815974} metrics. They approximate the grade level needed to comprehend the text and results indicate a more mature writing style in IIW human-authored and model generated outputs.}
\label{table:readability-stats}
\end{table*}

In addition to our human SxS comparisons, we use a suite of \textit{readability} metrics to quantify writing style differences between DCI, DOCCI, and IIW. We run heuristics based readability metrics over both human-authored and model-generated descriptions representing each style, and present the results in Table~\ref{table:readability-stats}. Each metric roughly estimates the level of education needed to understand a piece of written text using different units, \eg education years or grade-level. While they are proxy signals, a pattern across all can be seen as a clear indication of a more mature and articulate writing style for IIW in comparison with the other alternatives.  

\begin{figure*}[tb]
\centering
\begin{subfigure}{\linewidth}
\includegraphics[page=1, width=\linewidth]{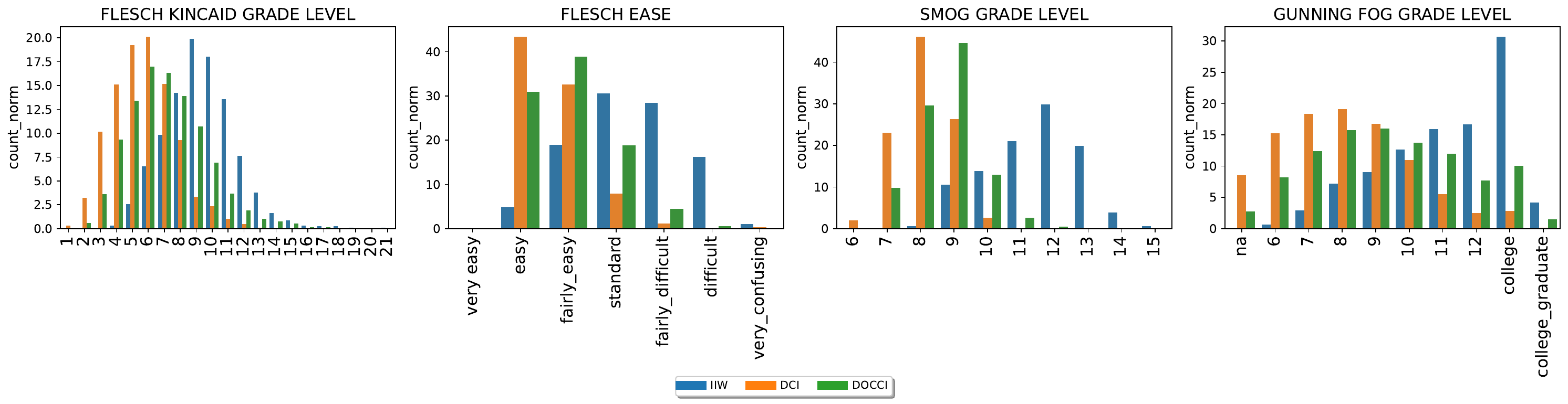}
\caption{Distribution on the Human Authored Datasets from DCI, DOCCI and IIW.}
\end{subfigure}
\begin{subfigure}{\linewidth}
\includegraphics[page=1, width=\linewidth]{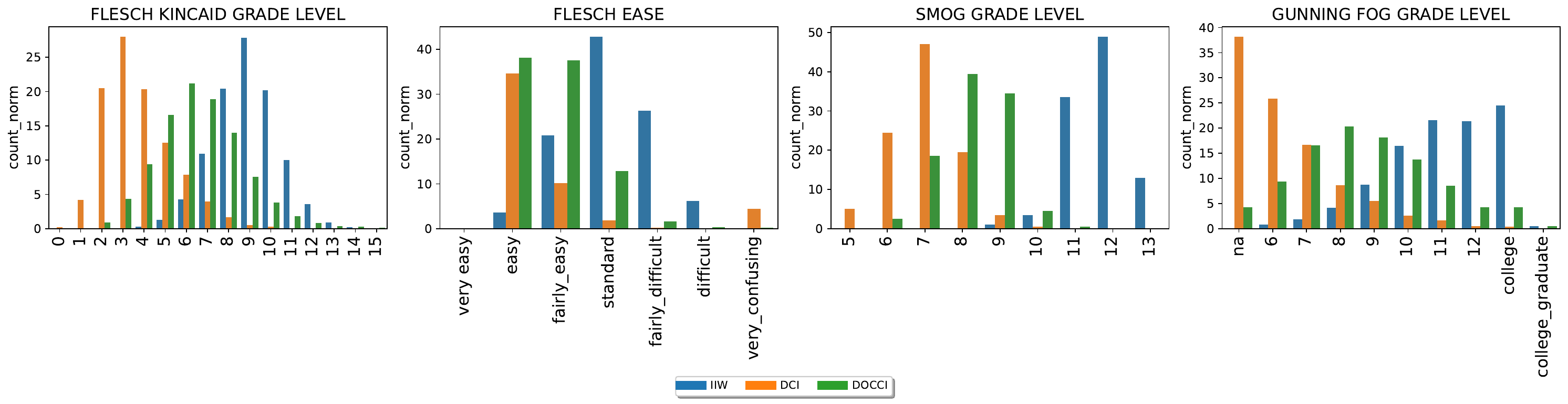}
\caption{Distribution on the Fine-tuned Model Generated Outputs from DCI, DOCCI and IIW.}
\end{subfigure}
\caption{Distribution-based Readability Metrics. We compare both human authored and model generated outputs from IIW and prior work to show the distribution of Education based units reflected in the writing style. IIW outputs from both the human annotators and the model produce a more mature style across the metrics.}
\label{fig:suplemental-readibility-fig}
\end{figure*}

For the metrics, we used spaCy~\cite{Honnibal_spaCy_Industrial-strength_Natural_2020} (v3.0.0rc2) to tokenize the text and the implementation in Github's \href{https://github.com/cdimascio/py-readability-metrics}{py-readability-metrics repo} (v1.4.1) to calculate the scores. We also include the readability metric distributions in Figure~\ref{fig:suplemental-readibility-fig}. The distributions further demonstrate a more mature writing style in both the IIW human-authored dataset and fine-tuned model generated outputs.

\setlength{\tabcolsep}{3pt}
\begin{table*}[tb]
\centering
\begin{tabular}{l|ccc|ccc|ccc}
\hline
\multirow{2}{*}{PaLI-ft} & \multicolumn{3}{c|}{DCI Test (112)} & \multicolumn{3}{c|}{DOCCI Test (5k)} & \multicolumn{3}{c}{IIW Test (445)} \\
\cline{2-10}
& bleu-4 & rouge-1 & rouge-2 & bleu-4 & rouge-1 & rouge-2 & bleu-4 & rouge-1 & rouge-2 \\
\hline
DCI & \textbf{4.97} & \textbf{35.38} & \textbf{12.70} & 5.24 & 39.55 & 12.95 & 2.30 & 31.70 & 8.58\\
DOCCI & 4.24 & 34.60 & 10.70 & \textbf{8.68} & \textbf{45.50} & \textbf{17.07} & 3.50 & 36.10 & 10.02\\
IIW & 3.02 & 31.59 & 8.02 & 4.60 & 38.10 & 10.06 & \textbf{5.66} & \textbf{38.57} & \textbf{11.73} \\
\hline
\end{tabular}
\label{tab:finetuning}
\caption{Cross Dataset Automatic Metric Evaluation of Fine-tuned Models.}
\label{table:cross-dataset-auto-metric-eval}
\end{table*}

\subsection{Side-by-Side (SxS) Evaluation Framework}
\label{sec:supplemental-evalutation-sxs}

\begin{figure*}[tb]
\centering
\begin{subfigure}{\linewidth}
\includegraphics[width=\linewidth]{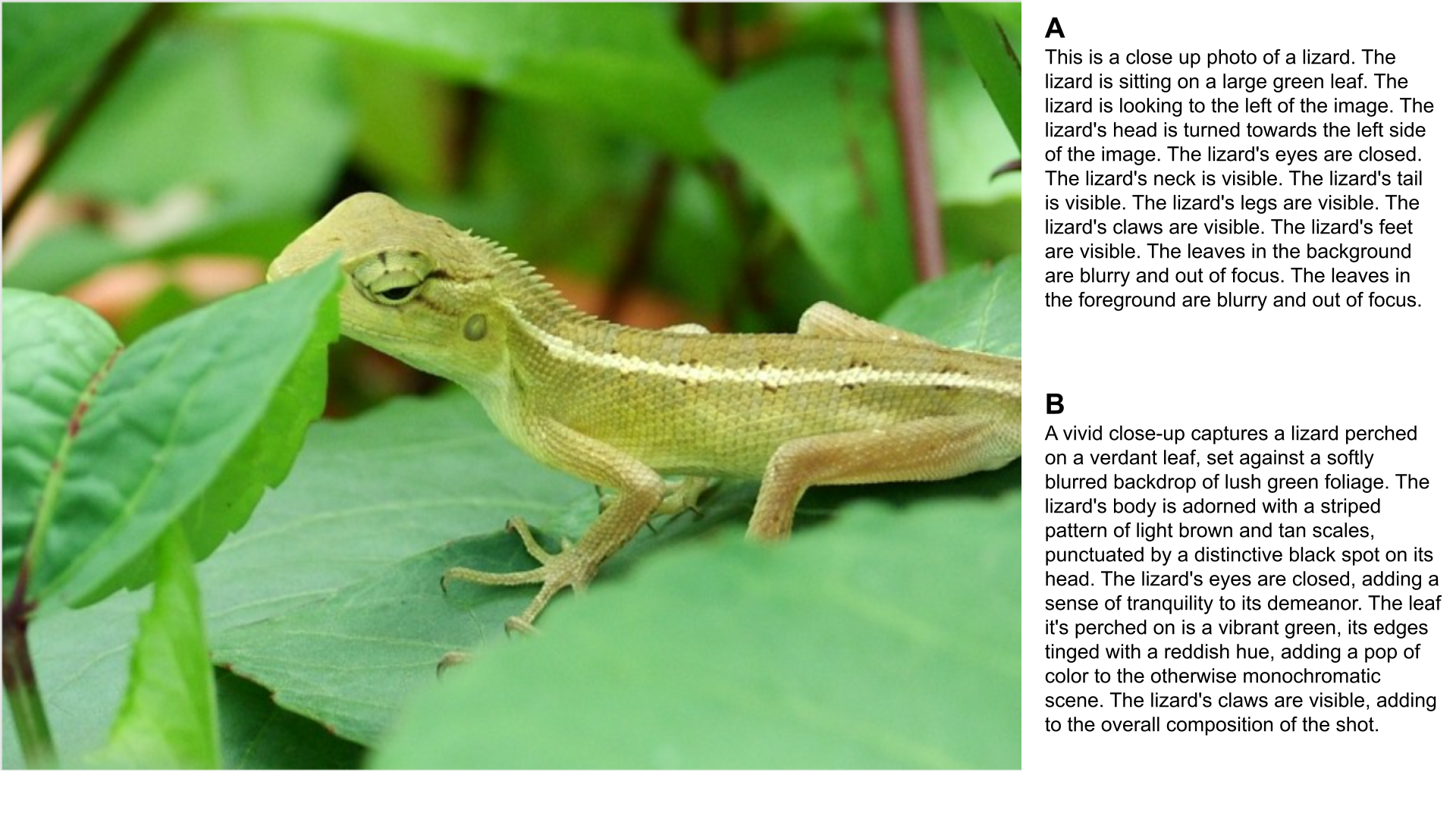}
\end{subfigure}
\caption{Human SxS Annotation UI. Annotators are shown the input image and two input image descriptions to evaluate side-by-side. The input descriptions could be from any combination of (human, model) sources. This information is not shared with the annotators and the sources are randomly flipped and marked as \textit{A} or \textit{B} to prevent any source or order based bias.}
\label{fig:suplemental-exp-human-sxs-ui-input}
\end{figure*}

\begin{figure*}[tb]
\centering
\begin{subfigure}{\linewidth}
\includegraphics[width=\linewidth]{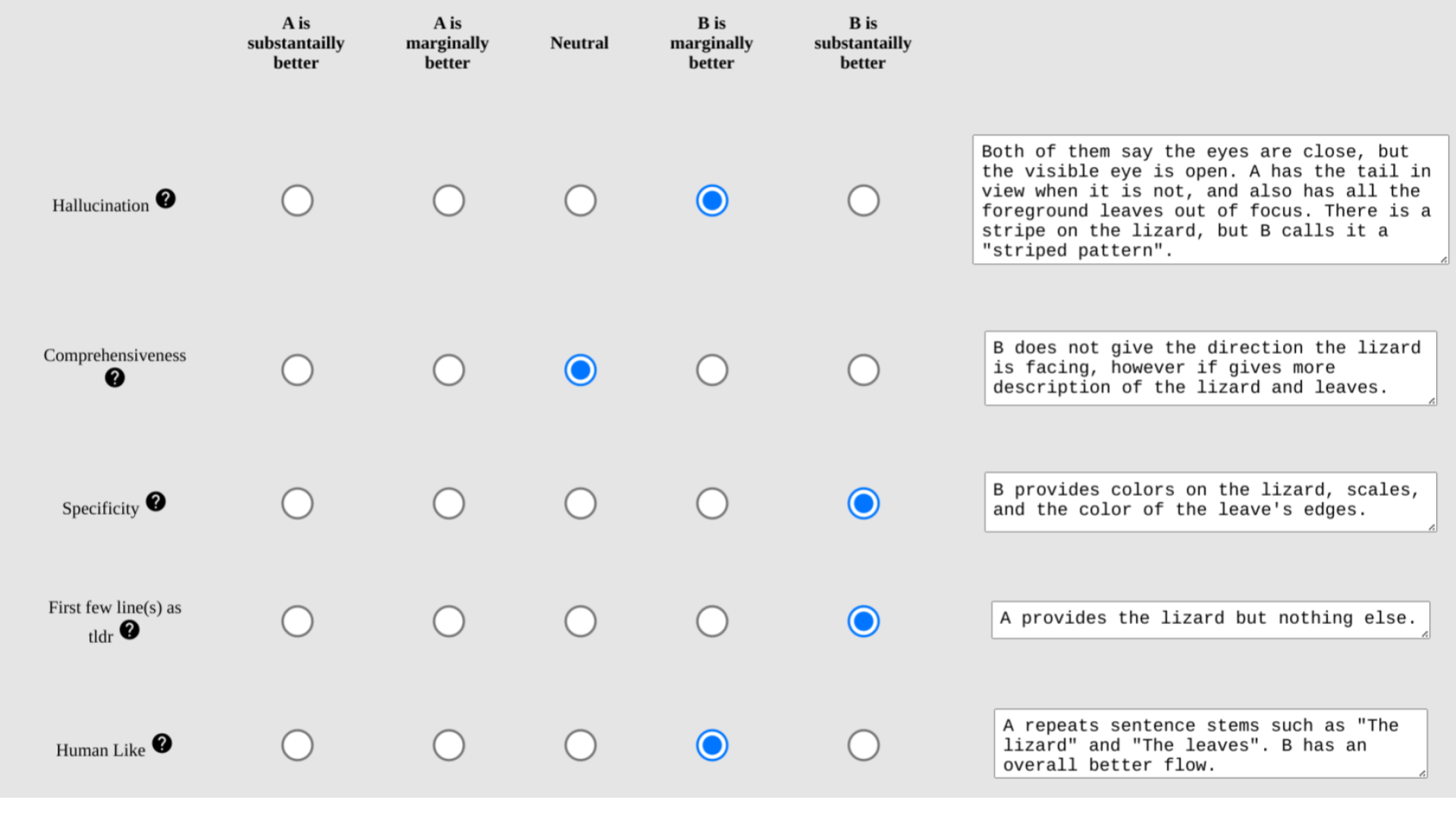}
\end{subfigure}
\caption{Human SxS Annotation UI responses for the input image and two image description pairs (see Figure~\ref{fig:suplemental-exp-human-sxs-ui-input}). The annotators respond to the 5 metrics independently on a 5 point scale. They are additionally asked to justify their choices which can be used to sanity check and perform quality sweeps.}
\label{fig:suplemental-exp-human-sxs-ui-annotation}
\end{figure*}

We demonstrate the Human SxS annotation UI to show the input (see Figure~\ref{fig:suplemental-exp-human-sxs-ui-input}) and the corresponding human responses (see Figure~\ref{fig:suplemental-exp-human-sxs-ui-annotation}) across the 5 metrics, each on a 5 point scale. The metrics are defined as:

\begin{itemize}
\item \textbf{Comprehensiveness:} The description should capture all of the important elements of the image, including objects, people, locations, actions, relationships between objects, \etc.
\item \textbf{Specificity:} The description should use precise and descriptive language to avoid vagueness and ambiguity. \Eg ``3 apples'' and ``Taj Mahal'' are more specific than ``some apples'' and ``a white marble structure,'' respectively.
\item \textbf{Hallucinations:} The description should be factually correct and avoid making assumptions or interpretations that are not visually supported by the image.
\item \textbf{First few line(s) as tldr:} The first few line(s) should paint a high level picture of what to expect in the image and create a succinct summary.
\item \textbf{Human-Like:} The descriptions should feel as if an educated person wrote them and should be free from artifacts hinting that a machine generated them (\eg stuttering, repeating facts, fragmented chain of thought, \etc).
\end{itemize}

The 5 metrics are defined to capture 3 broad umbrella metrics of precision, recall and writing-style. An \textit{overall metric} score can further be computed by taking an average of the 3 umbrella metrics. Each can be defined as follows:
\begin{equation*}
\begin{aligned}
\texttt{Recall} &= avg(\text{Comprehens.}, \text{Specific.})\\
\texttt{Precision} &= \text{Hallucination}\\
\texttt{Writing Style} &= avg(\text{TLDR}, \text{Human Like})\\
\texttt{Overall} &= avg(\text{Rec.}, \text{Prec.}, \text{Writing Sty.})
\end{aligned}
\end{equation*}

 \begin{table*}[tb]
\centering
\begin{tabular}{l|ccc|ccc|ccc}
\hline
\multirow{2}{*}{PaLI-ft} &\multicolumn{3}{c|}{DCI Test (112)}&\multicolumn{3}{c|}{DOCCI Test (5k)}&\multicolumn{3}{c}{IIW Test (445)}\\
& CIDEr &BERT & BLEURT & CIDEr &BERT & BLEURT & CIDEr &BERT & BLEURT\\
\hline
DCI & 4.57 & \textbf{0.60} & 0.41 & 4.71 & 0.61 & 0.42 & 0.75 & 0.56 & 0.40 \\
DOCCI & \textbf{4.91} & 0.58 & 0.39 & \textbf{11.09} & \textbf{0.65} & 0.45 & 2.40 & 0.59 & 0.41 \\
IIW & 1.87 & 0.56 & 0.41 & 4.52 & 0.59 & \textbf{0.46} & \textbf{4.04} & \textbf{0.61} & 0.45 \\
IIW Comb. & 0.61 & 0.56 & \textbf{0.43} & 4.15 & 0.59 & \textbf{0.46} & 1.77 & 0.60 & \textbf{0.46} \\
\hline
\end{tabular}
\caption{Additional Automatic Metric Results. We report CIDEr, BERTScore (referred to as BERT in table due to space), and BLEURT metrics for all fine-tuned models. We compare DCI, DOCCI, IIW, and IIW Comb. (Combined).}
\label{table:supplemental-iiw-finetuning-more-metrics}
\end{table*}

\subsection{Additional Automatic Metrics}
\label{sec:textsimmetrics}
We include evaluations of model-generated outputs with automated text similarity metrics for completeness, but note that common text similarity metrics are ill-suited for long texts and more recent image-text metrics are often length limited. We report these results simply to emphasize the limitations of these metrics when measuring the quality of hyper-detailed image descriptions. Using standard automatic metrics, Table~\ref{table:cross-dataset-auto-metric-eval} illustrates how fine-tuned models largely perform better in replicating their own style. 

In addition to reporting BLEU-4, ROUGE-1, and ROUGE-2 automatic metrics, we include CIDEr~\cite{vedantam2015cider}, BERTScore~\cite{bertscore}, and BLEURT~\cite{pu2021learning} metrics in Table~\ref{table:supplemental-iiw-finetuning-more-metrics}. We include BERTScore and BLEURT as they are newer, model-based metrics which have been shown to correlate more closely with human judgements. CIDEr, like BLEU and ROUGE metrics are not limited by sequence length. BERTScore and BLEURT have a maximum sequence length of 512 (we specifically use the ``wwm\_cased\_L-24\_H-1024\_A-16'' BERT checkpoint and the latest BLEURT-20 model), but for our descriptions, they likely fit under this maximum length, with only outliers being truncated.

CIDEr and BERTScore generally show the same trend of each fine-tuned model performing best on the same test domain (\ie, DCI fine-tuned models perform best on DCI test set, DOCCI models perform best on DOCCI test set, and so on). One anomaly occurs with CIDEr on the DCI test set, where PaLI models fine-tuned with DOCCI slightly outperform the DCI trained model (4.91 versus 4.57). Due to how low the metric values are, these differences may not be significant. When evaluating the DCI, DOCCI, and IIW test sets with BLEURT, we instead find a slight preference for IIW models. Across all three datasets, BLEURT shows PaLI-IIW variants perform better or similarly to the same-domain test set. Thus, newer metrics may reveal IIW fine-tuned models generalize better than models fine-tuned on other datasets.

\setlength{\tabcolsep}{1.5pt}
\begin{table*}[tb]
\centering
\begin{tabular}{l|ccc|ccc|ccc}
\hline
\multirow{2}{*}{PaLI-ft} & \multicolumn{3}{c|}{DCI Test (112)} & \multicolumn{3}{c|}{DOCCI Test (5k)} & \multicolumn{3}{c}{IIW Test (445)} \\
& bleu-4 & rouge-1 & rouge-2 & bleu-4 & rouge-1 & rouge-2 & bleu-4 & rouge-1 & rouge-2 \\
\hline
IIW & \textbf{3.02} & \textbf{31.59} & \textbf{8.02} & 4.60 & 38.10 & 10.06 & \textbf{5.66} & \textbf{38.57} & \textbf{11.73} \\
IIW Combined & 2.95 & 30.63 & 7.30 & \textbf{4.76} & \textbf{38.25} & \textbf{10.48} & 5.40 & 37.64 & 11.62 \\
\hline
\end{tabular}
\caption{Ablation Results Comparing IIW Variants on Automatic Metrics.}
\label{table:supplemental-iiw-finetuning-ablations}
\end{table*}
\subsection{IIW Fine-tuned Model Ablations}
As an IIW ablation study, we fine-tune a separate PaLI-5B model, \textit{IIW-Combined}, using all the data from Task 1 and Task 2 as a mixture of 7 training tasks, defined in Appendix~\ref{sec:supplemental-tasks}.
Table \ref{table:supplemental-iiw-finetuning-more-metrics} and \ref{table:supplemental-iiw-finetuning-ablations} show that this has no clear significant gains on Task-2's final description eval set. This currently remains a less explored area and we aim to investigate this in future work to further improve the model on Task-2 evaluations.\setlength{\tabcolsep}{1.25pt}

\subsection{Reconstructing Images with IIW Descriptions}
\label{subsec:addtnt2i}
\begin{figure*}[tb]
\centering
\begin{subfigure}{\linewidth}
\includegraphics[width=\linewidth]{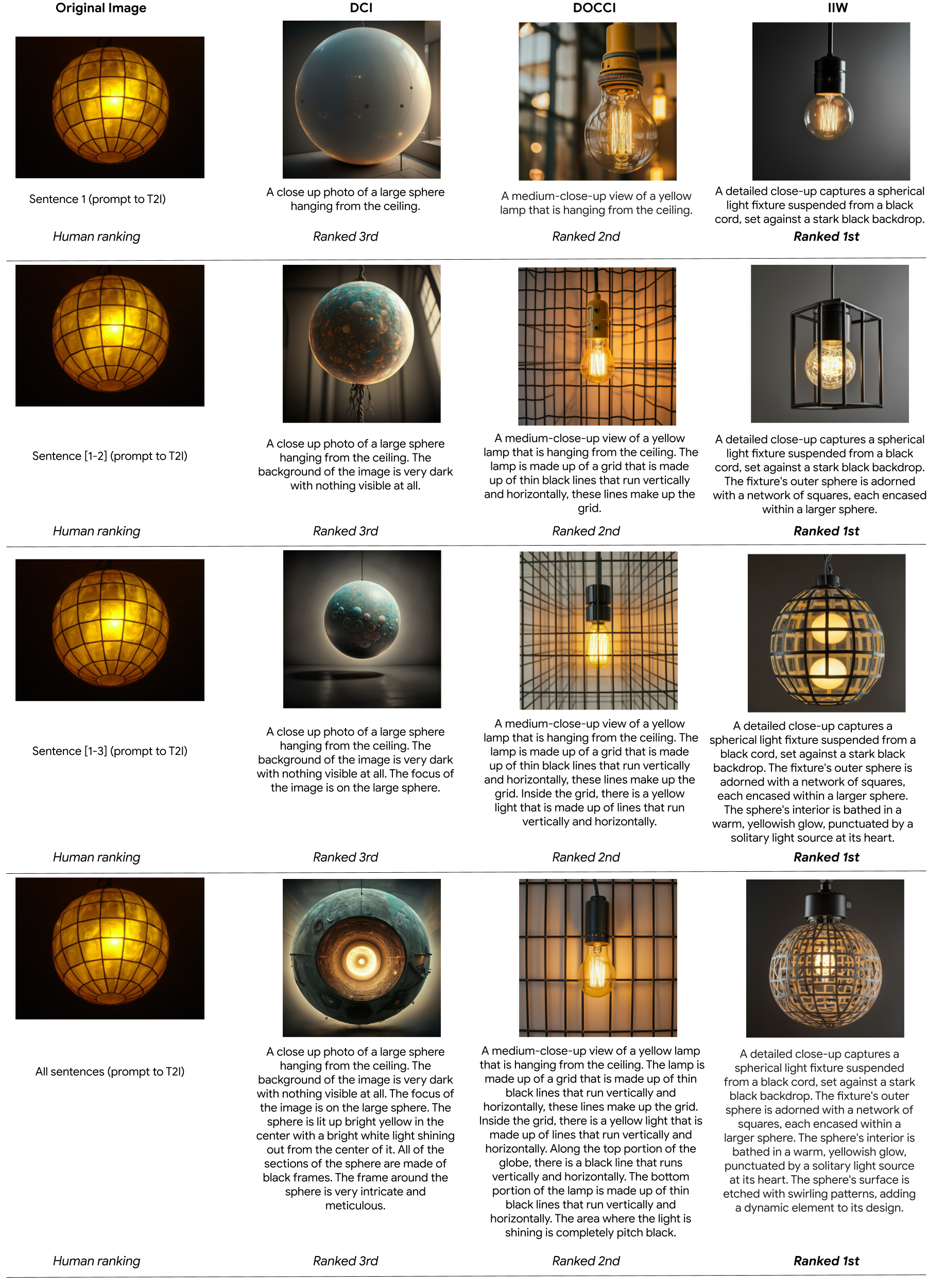}
\label{fig:ImageInWords_Figures-supp-reconstruction}
\end{subfigure}
\caption{T2I Outputs and Human Ranking Evaluations. We show example T2I results where the first sentence, first two sentences, ..., all the sentences of the image descriptions from DCI, DOCCI and IIW models are fed sequentially as inputs, \ie, at each step an additional sentence chunk is fed to the T2I model.}
\label{fig:supp-reconstruction}
\end{figure*}

\begin{figure*}[tb]
\centering
\begin{subfigure}{\linewidth}
\centering
\includegraphics[width=\linewidth]{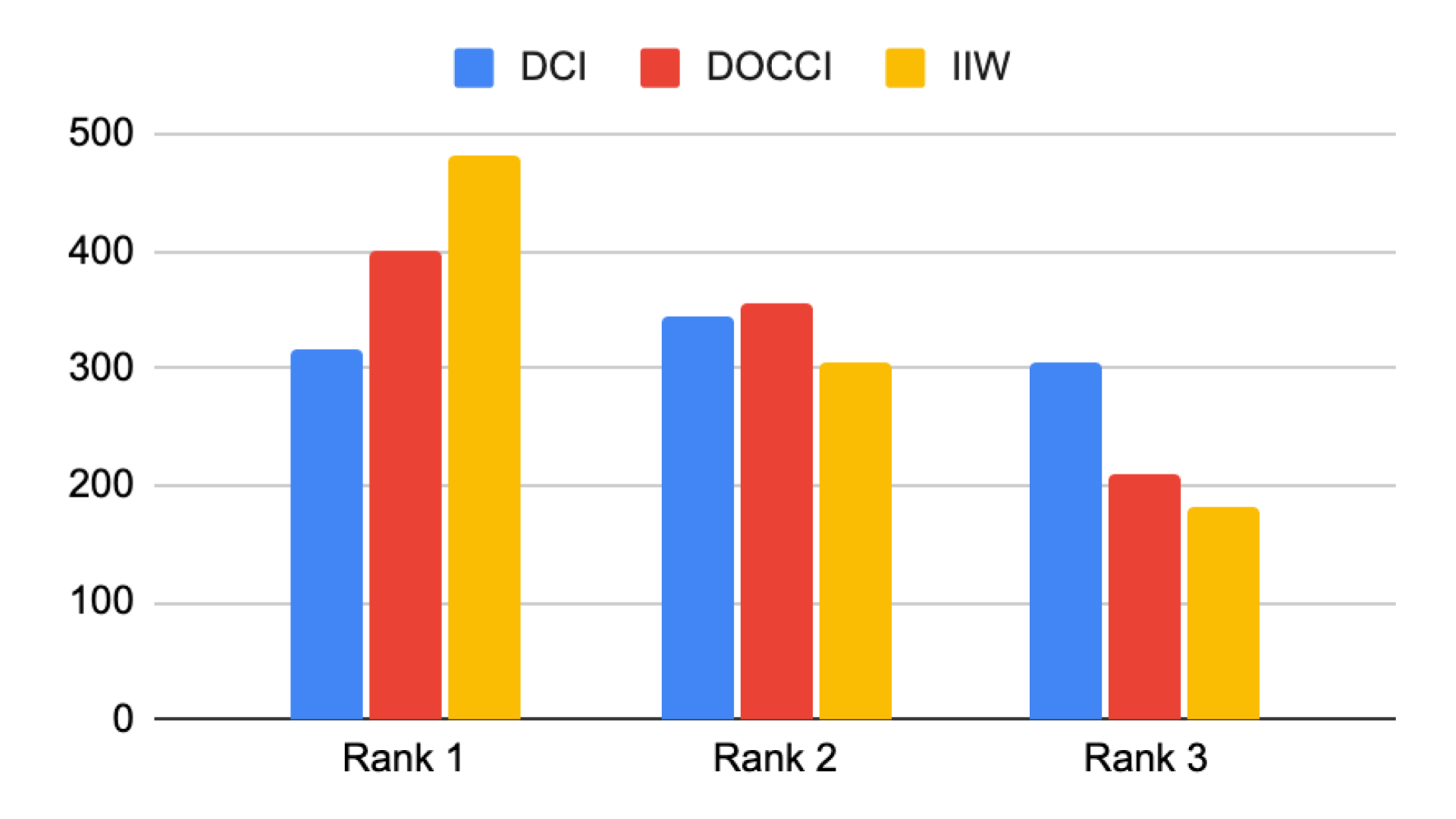}
\caption{Reconstruction Rank Counts across Inputs over All Cumulative Sentence Chunks.}
\end{subfigure}
\begin{subfigure}{\linewidth}
\includegraphics[width=\linewidth]{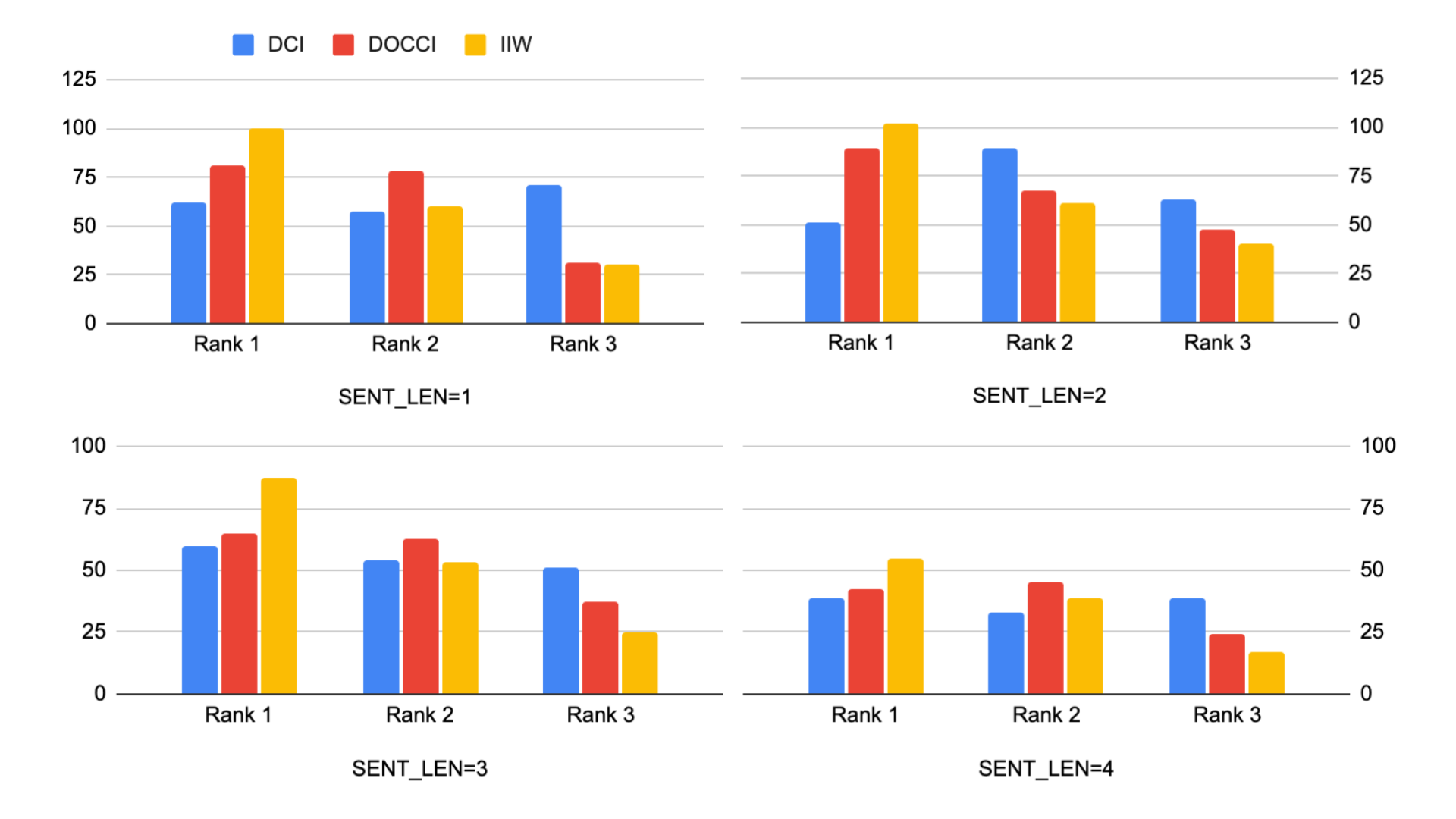}
\caption{Reconstruction Rank Counts across Inputs of Specific Cumulative Sentence Chunks.}
\end{subfigure}
\caption{T2I Human Rank Distributions. We illustrate bar plots for the image reconstruction evaluation results using image descriptions from fine-tuned PaLI-5B models on three datasets (DCI, DOCCI, IIW). Images reconstructed from IIW descriptions are consistently ranked better than other descriptions.}
\label{fig:suplemental-t2i-charts}
\end{figure*}

For reconstructing images sentence-by-sentence, we fed the T2I model the \textit{first sentence}, \textit{first two sentences}, \textit{first three sentences}, \etc. as prompts from each of the three datasets (DCI, DOCCI and IIW). Figure~\ref{fig:supp-reconstruction} showcases the prompts and the T2I model outputs from three descriptions along with the original image. 

We then asked human annotators to rank the generated images by how similar they are to the original image. The image most similar to the original image is ranked number 1. We allowed generated images to be ranked the same if they are very similar. Figure~\ref{fig:suplemental-t2i-charts}(a) shows the reconstruction rank counts for all the sentence counts and Figure~\ref{fig:suplemental-t2i-charts}(b) shows the rank counts when we use sentence 1, sentence 1 and 2, sentence 1, 2 and 3, and sentence 1, 2, 3, and 4. Sentences from IIW descriptions are ranked first much more frequently than sentences from DCI and DOCCI descriptions. Specifically, for the first sentence, the difference is most notable, supporting our claim that IIW descriptions are higher quality earlier on and IIW first sentences are designed to capture a TLDR.

\subsection{Compositional Reasoning with IIW Descriptions}
\label{subsec:appendixllmvqa}
In our downstream evaluation of ARO, SVO-Probes, and Winoground compositional reasoning benchmarks with IIW descriptions, we formulate a new LLM-only method of evaluation. We prompt a LLM (\eg, PaLM 2) to determine which is the true matching caption given the generated image description and the image caption options to select from.
We define the LLM prompt which includes an image description as:

\begin{quote}
``Given the following image description and image caption options, choose the most likely OPTION number :

IMAGE-DESCRIPTION : <DESCRIPTION>

OPTIONS :
<CHOICES>

RESPONSE : 
''
\end{quote}

where we fill in the <DESCRIPTION> from each VLM description model (\eg, either our IIW fine-tuned model, InstructBLIP or LLaVA) and the list of <CHOICES> are from the corresponding evaluation dataset, respectively. Choices are enumerated in a list-like fashion, and we ask the model to generate the number of the most likely caption. 

We define a different prompt for the language bias baseline, which serves as a sanity check that the image/image description is truly needed for these datasets. It provides a lower bound for comparison, too. While the prompt is different as we do not input any image description, we try to make it as similar as possible to the above image description based prompt. We set the language bias prompt to:

\begin{quote}
``Given the following image caption options, choose the most likely OPTION number :

OPTIONS :
<CHOICES>

RESPONSE : ''
\end{quote}

where <CHOICES> are filled in in the same format as previously described.

Importantly, when filling in the caption choices, we deterministically swap the index of the ``answer,'' \ie, the true matching caption, among the choices list in the prompt. This is done to ensure an equal distribution and reduce any order bias (\eg, a LLM may be more prone to believing the first option is the correct option).

To obtain the image description which is then fed into the LLM, we prompt our fine-tuned models with ``Generate a detailed image description.'' For the  InstructBLIP and LLaVA models, we define similar prompts given the prompts used in their published papers papers: ``Write a long and detailed description for the photo.'' and ``Provide a detailed description of the given image'' for InstructBLIP and LLaVA, respectively.

We process the LLM outputs as classes, (\eg, when choosing between image caption choices [1] and [2], LLM responses are `1' or `2') and calculate accuracy with respect to the true image caption class. If the LLM does not produce a valid class, it's considered an incorrect prediction. Note that this task set up is different from how VLM models are typically evaluated on these reasoning datasets: prior work considers a sample to be correctly reasoned about if the image-text similarity of the true image caption is higher than the image-text similarity of the incorrect image caption. Due to the long length of our descriptions, we cannot compute image-text similarity reasonably with models like CLIP without significantly truncating our image descriptions. In future work, once input length limitations are mitigated, dual-encoder VLMs like CLIP can be fine-tuned with our rich data, which will help to improve VLM reasoning.

Note that ARO and Winoground datasets are built with positive and negative captions for each image. SVO-Probes differs in that it originally contained a positive and negative \textit{image} for each positive caption. For our experiments, we need a true and false caption associated with an image. A large portion ($\sim$90\%) of the SVO-Probes negative images also serve as separate samples (where they are considered positive images, with associated captions). Thus, we can pull these captions to serve as the negative caption for the original sample.

For the remaining $\sim$10\%, we use the negative triplet (the S, V, O triplet specifying the subject, object, and verb, with one of them being modified) to automatically flip the negative S, V, or O in the positive caption. Ten of these samples did not have negative triplets in the dataset, so they were removed. Lastly, there were 114 samples with positive captions not containing the S, V, or O that needed to be swapped to form the negative caption. This happens as a result of SVO triplets containing root forms of the words, which were not spelled the same way in the caption. For example, an SVO may be ``man,lie,beach'' with the caption stating ``A man lying on a beach.'' Due to the verb tense differences, it would require additional processing to match ``lie'' to ``lying.'' We remove these edge cases for simplicity.

\begin{table*}[tb]
\centering
\begin{tabular}{lcccc}
\hline
\multirow{2}{*}{\shortstack[p]{Image Description Model}} & \multicolumn{2}{c}{ARO} & \multirow{2}{*}{SVO-Probes} & \multirow{2}{*}{Winoground} \\
& VG-A & VG-R & & \\
\hline
None (Language Bias Baseline) & 56.50 & 59.94 & 50.71 & 49.88 \\
InstructBLIP-Vicuna-\textit{7B} & 83.99 & 62.73 & \textbf{89.35} & 65.25  \\
LLaVA-V1.5-\textit{7B} & 84.80 & 63.71 & 87.89 & 63.38\\
PaLI-3 + DCI \textit{5B} & 88.19 & 66.47 & 86.50 & 64.62 \\
PaLI-3 + DOCCI \textit{5B} & 89.70 & \textbf{68.85} & 88.73 & \textbf{69.50} \\
PaLI-3 + IIW \textit{5B} & \textbf{90.37} & 66.19 & 88.66 & 69.38\\
PaLI-3 + IIW Combined \textit{5B} & 89.46 & 64.88 & 87.78 & 66.88\\
\hline
\end{tabular}
\caption{VL Compositional Reasoning Accuracy with Image Descriptions. We evaluate whether rich descriptions can distinguish the true matching image caption in ARO~\cite{yuksekgonul2023visionlanguage}, SVO-Probes~\cite{hendricks2021probing}, and Winoground~\cite{thrush2022winoground} datasets. The COCO and Flickr30k Order subsets of ARO are not reported due to a very high language bias baseline of 98\%.}
\label{table:llm-vqa-full}
\end{table*}

Finally, we include more vision language compositional reasoning results with different PaLI fine-tuned models in Table~\ref{table:llm-vqa-full}. Here we additionally include the models fine-tuned with DCI and DOCCI datasets. The IIW descriptions still result in highest reasoning accuracy for ARO VG-A and are comparable with DOCCI on Winoground. Trends also stay the same with SVO-Probes, with DOCCI performing similarity to IIW, but InstructBLIP performing slightly better (by less than 1 accuracy point). Finally, we find that DOCCI performs best on VG-R, which might be result of its dataset being designed to explicitly contain connected and contrasting images, which might more frequently capture similar images that only differ by the visual relationship between objects.

While performance differences between DCI, DOCCI, and IIW are smaller, this could be an artifact of the reasoning datasets; ARO, SVO-Probes, and Winoground are all built upon short caption datasets, so the utility and quality differences between DCI, DOCCI, and IIW are not fully captured by these probing datasets. 

\section{Enriching Image Caption Datasets}
\label{sec:enriched}
\setlength{\tabcolsep}{3.5pt}
\begin{table*}[tb]
\centering
\begin{tabular}{l|c|c|c|c|c|c|c|c}
\hline
\multirow{2}{*}{Dataset} & Sample &  Tokens & Tokens & Sentences & NN & ADJ & ADV & VB \\ 
\cline{3-9}
& Count  & / Sent. & \multicolumn{6}{c}{/ Desc.}\\
\hline 
LocNar~\cite{PontTuset_eccv2020} & \multirow{2}{*}{1000} & 14.35 & 30.56 & 2.12  & 8.02 & 1.09 & 0.16 & 2.39\\
\hspace{1mm} IIW Enriched &  & \textbf{22.19} & \textbf{128.87} & \textbf{5.80} & \textbf{32.37} & \textbf{16.02} & \textbf{1.82} & \textbf{11.44}\\
XM3600~\cite{thapliyal2022crossmodal3600} &  \multirow{2}{*}{1000} & 10.40 & 10.40 & 1.00 & 3.45 & 1.08 & 0.04 & 0.61 \\
\hspace{1mm} IIW Enriched &  & \textbf{22.25} & \textbf{130.56} & \textbf{5.86} & \textbf{33.18} & \textbf{15.82} & \textbf{1.72} & \textbf{11.87}\\
\hline 
\end{tabular}
\caption{Dataset Statistics Comparing ImageInWords (IIW) Descriptions of Prior Work to their Original Annotations. We include the number of samples (\ie, subset of captions/descriptions that we enrich) and the average number of tokens, sentences, nouns (NN), adjectives (ADJ), adverbs (ADV), and verbs (VB). Language statistics are averages reported per description unless otherwise noted.}
\label{table:enrichlangcompare}
\end{table*}

As discussed in the main paper, we enrich 1k samples from two existing image caption datasets, namely, Localized Narratives and CrossModal (XM) 3600, with new image descriptions generated by IIW fine-tuned models. The goal of releasing these enriched versions is to provide longer, hyper-detailed image descriptions that can be used for evaluation purposes in future work. The enriched versions not only allow for finer-grained, full coverage evaluations of the content in images (via new metrics or probing datasets), but also may enable autorater models which learn from the precision and recall errors in the generated descriptions.

In Table~\ref{table:enrichlangcompare}, we report the language statistics on the original 1k samples from each dataset and the enriched versions. It is clear that the IIW descriptions are significantly longer and richer, as we have higher counts of tokens, sentences, and each part of speech.

\section{Percentages Reported in the Main Paper}
We re-quote and define all analysis percentages reported in the main paper for clarity on how they were calculated in Tables~\ref{tab:calculations}-\ref{tab:calculations3}. The reference location is defined by the section, paragraph, and line it appeared in. We only include paragraph number for multi-paragraph sections, and only include line number if the same percentage occurs more than once within a paragraph. For example, ``S4.3 P2 L3'' means Section 4, Paragraph 2, Line 3. Most percentages were rounded to the nearest point in the main paper. 

\begin{table*}[t]
\centering
\begin{tabular}{c|>{\raggedright\arraybackslash}p{1.6cm}|>{\raggedright\arraybackslash}p{8.25cm}}
\hline
Percent & Reference & Equation and Explanation \\
\hline
+66\% & Abstract, Intro P5, Conclusion & Average difference of IIW preference vs. other dataset preference, averaged over DCI and DOCCI datasets and averaged over the five metrics corresponding to (comprehensiveness, specificity, hallucinations, tldr, human-likeness). Differences of IIW marginally and substantially better - other dataset marginally and substantially better for (comprehensiveness, specificity, hallucinations, tldr, human-likeness) metrics from Table 2 correspond to DCI (61, 80, 42, 91, 82) and DOCCI (42, 82, 35, 79, 68). The final average preference over the five metrics and two datasets is 66.2\%. \\
\hline
+48\% & Abstract, Intro P5 & Average difference of IIW preference vs. GPT-4V outputs, averaged over the five metrics corresponding to (comprehensiveness, specificity, hallucinations, tldr, human-likeness). Differences of IIW marginally and substantially better - GPT-4V marginally and substantially better for (comprehensiveness, specificity, hallucinations, tldr, human-likeness) metrics from Table 3 correspond to (35, 53, 59, 70, 21). The final average preference over the five metrics is 47.6\%. \\
\hline
+31\% & Abstract, Intro P5, S5.1 P1, Conclusion & 
Average difference of IIW model output preference vs. other fine-tuned model output preference, averaged over DCI and DOCCI fine-tuned models and averaged over the five metrics corresponding to (comprehensiveness, specificity, hallucinations, tldr, human-likeness). Differences of IIW marginally and substantially better - other dataset marginally and substantially better for (comprehensiveness, specificity, hallucinations, tldr, human-likeness) metrics from Table 3 correspond to DCI (42, 54, -9, 51, 57) and DOCCI (4, 37, -7, 57, 23). The final average preference over the five metrics and two datasets is 30.9\%. \\
\hline 
20\% more & S3.2 P6 & The median increase in token count from annotation round 1 to round 3: (205-170)/170 = 20\%. \\
\hline
30\% less & S3.2 P6 & The median decrease in time spent annotating from round 1 to round 3 compared to if three individual round 1s occurred: ((800*3)-(800+600+300))/(800*3) = 30\%. \\
\hline
+61\% & S4.1 P1 & The amount IIW is more comprehensive than DCI in Table 2: (30+41) - (3+7) = 61\%. \\
\hline
+42\% & S4.1 P1 L4 & The amount IIW is more comprehensive than DOCCI in Table 2: (33+19) - (4+6) = 42\%.\\
\hline
+80\% & S4.1 P1 L5 & The amount IIW is more specific than DCI in Table 2: (20+68) - (5+3) = 80\%. \\
\hline
\end{tabular}
\caption{Percentages from the Main Text. We reference each percentage and define how they were calculated for clarity.}
\label{tab:calculations}
\end{table*}

\begin{table*}[t]
\centering
\begin{tabular}{c|>{\raggedright\arraybackslash}p{1.6cm}|>{\raggedright\arraybackslash}p{8.25cm}}
\hline
Percent & Reference & Equation and Explanation \\
\hline
+82\% & S4.1 P1 L5 & The amount IIW is more specific than DOCCI in Table 2: (22+65) - (3+2) = 82\%. \\
\hline
42\% & S4.1 P1 L5 & The amount IIW contains fewer hallucinations than DCI in Table 2: (32+15) - (2+3) = 42\%. \\
\hline
35\% & S4.1 P1 L6 & The amount IIW contains fewer hallucinations than DOCCI in Table 2: (34+13) - (0+12) = 35\%. \\
\hline
+91\% & S4.1 P1 L6 & The amount IIW contains better TLDR than DCI in Table 2: (20+74) - (3+0) = 91\%.\\
\hline
+79\% & S4.1 P1 L7 & The amount IIW contains better TLDR than DOCCI in Table 2: (30+54) - (1+4) = 79\%. \\
\hline
+82\% & S4.1 P1 L7 & The amount IIW is more human-like than DCI in Table 2: (25+59) - (1+1) = 82\%.  \\
\hline
+68\% & S4.1 P1 L8 & The amount IIW is more human-like than DOCCI in Table 2: (46+23) - (1+0) = 68\%. \\
\hline
+35\% & S4.1 P2 & The amount IIW is more comprehensive than GPT-4V outputs in Table 3: (29+19)-(3+10) = 35\%. \\
\hline
+53\% & S4.1 P2 & The amount IIW is more specific than GPT-4V outputs in Table 3: (35+34) - (6+10) = 53\%. \\
\hline
+59\% & S4.1 P2 & The amount IIW is contains fewer hallucinations than GPT-4V outputs in Table 5: (34+31) - (0+6) = 59\%. \\
\hline
+70\% & S4.1 P2 & The amount IIW contains better TLDR than GPT-4V outputs in Table 3: (47+34) - (5+6) = 70\%. \\
\hline
+21\% & S4.1 P2 & The amount IIW is more human-like than GPT-4V outputs in Table 3: (27+13) - (6+13) = 21\%. \\
\hline
+42\% & S5.1 P1 & The amount IIW is more comprehensive than DCI in Table 3: (32+27) - (7+10) = 42\%. \\
\hline
+4\% & S5.1 P1 & The amount IIW is more comprehensive than DOCCI in Table 3: (26+5) - (5+22) = 4\%. \\
\hline
+54\% & S5.1 P1 & The amount IIW is more specific than DCI in Table 3: (24+46) - (6+10) = 54\%. \\
\hline
+37\% & S5.1 P1 & The amount IIW is more specific than DOCCI in Table 3: (33+24) - (6+14) = 37\%. \\
\hline
\end{tabular}
\caption{Percentages from the Main Text. We reference each percentage and define how they were calculated for clarity.}
\label{tab:calculations2}
\end{table*}

\begin{table*}[t]
\centering
\begin{tabular}{c|>{\raggedright\arraybackslash}p{1.6cm}|>{\raggedright\arraybackslash}p{8.5cm}}
\hline
Percent & Reference & Equation and Explanation \\
\hline
+51\% & S5.1 P1 & The amount IIW contains better TLDR than DCI in Table 3: (30+41) - (9+11) = 51\%. \\
\hline
+57\% & S5.1 P1 & The amount IIW contains better TLDR than DOCCI in Table 3: (42+28) - (6+7) = 57\%. \\
\hline
+55\% & S5.1 P1 & The amount IIW is more human-like than DCI in Table 3: (32+39) - (11+5) = 55\%. \\
\hline
+23\% & S5.1 P1 & The amount IIW is more human-like than DOCCI in Table 3: (27+14) - (6+12) = 23\%. \\
\hline
-9\% & S5.1 P1 & The amount IIW contains fewer hallucinations than DCI in Table 3: (11+13) - (12+21) = -9\%.  \\
\hline
-7\% & S5.1 P1 & The amount IIW contains fewer hallucinations than DOCCI in Table 3: (21+6) - (9+25) = -7\%. \\
\hline
34\% & S5.3 P4 & The accuracy improvement on VG-A from using IIW over the language bias baseline: (90.37) - (56.50) = 33.87\%. \\
\hline
6\% & S5.3 P4 & The accuracy improvement on VG-R from using IIW over the language bias baseline: (66.19) - (59.94) = 6.25\%. \\
\hline
20\% & S5.3 P4 & The accuracy improvement on Winoground from using IIW over the language bias baseline: (69.38) - (49.88) = 19.5\%.\\
\hline
6\% & Abstract, S5.3 P4, Conclusion & The accuracy improvement on VG-A from using IIW over the next best baseline LLaVA: (90.37) - (84.80) = 5.57\%.\\
\hline
2\% & S5.3 P4 & The accuracy improvement on VG-R from using IIW over the next best baseline LLaVA: (66.19) - (63.71) = 2.48\%. \\
\hline
4\% & S5.3 P4 & The accuracy improvement on Winoground from using IIW over the next best baseline InstructBLIP: (69.38) - (65.25) = 4.13\%. \\
\hline
\end{tabular}
\caption{Percentages from the Main Text. We reference each percentage and define how they were calculated for clarity.}
\label{tab:calculations3}
\end{table*}

\end{document}